\documentclass[10pt,journal,compsoc]{IEEEtran}
\ifCLASSOPTIONcompsoc
  \usepackage[nocompress]{cite}
\else
  \usepackage{cite}
\fi

\usepackage{cite}
\usepackage{graphicx} 
\usepackage{float} 
\usepackage{subfigure} 
\usepackage{amssymb}
\usepackage{amsmath}
\usepackage{multirow}
\usepackage{color}
\usepackage{bm}
\usepackage[T1]{fontenc}
\usepackage[colorlinks,linkcolor=black]{hyperref}
\begin{document}
%
\title{Multi-Granularity Regularized Re-Balancing for Class Incremental Learning}
%
%
%
%

\author{Huitong~Chen,
        Yu~Wang,
        and~Qinghua~Hu, \IEEEmembership{Senior~Member,~IEEE}
\IEEEcompsocitemizethanks{
\IEEEcompsocthanksitem Huitong Chen, Yu Wang, and Qinghua Hu are with the College of Intelligence and Computing, Tianjin University, Tianjin 300350, China and Key Laboratory for Machine Learning of Tianjin, China. Yu Wang and Qinghua Hu are also with Haihe Laboratory of Information Technology Application Innovation, Tianjin, China. Corresponding author: Yu Wang, E-mail: wangyu\_@tju.edu.cn.
\IEEEcompsocthanksitem This work was supported in part by the National Natural Science Foundation of China under Grants 62106174, 61732011, and 61925602, and in part by the China Postdoctoral Science Foundation under Grants 2021TQ0242 and 2021M690118.
\protect}
}
%
%

\markboth{IEEE Transactions on Knowledge and Data Engineering}
{Shell \MakeLowercase{\textit{et al.}}: Bare Demo of IEEEtran.cls for Computer Society Journals}
%


\IEEEtitleabstractindextext{%
\begin{abstract}
Deep learning models suffer from catastrophic forgetting when learning new tasks incrementally. Incremental learning has been proposed to retain the knowledge of old classes while learning to identify new classes. A typical approach is to use a few exemplars to avoid forgetting old knowledge. In such a scenario, data imbalance between old and new classes is a key issue that leads to performance degradation of the model. Several strategies have been designed to rectify the bias towards the new classes due to data imbalance. However, they heavily rely on the assumptions of the bias relation between old and new classes. Therefore, they are not suitable for complex real-world applications. In this study, we propose an assumption-agnostic method, Multi-Granularity Regularized re-Balancing (MGRB), to address this problem. Re-balancing methods are used to alleviate the influence of data imbalance; however, we empirically discover that they would under-fit new classes. To this end, we further design a novel multi-granularity regularization term that enables the model to consider the correlations of classes in addition to re-balancing the data. A class hierarchy is first constructed by grouping the semantically or visually similar classes. The multi-granularity regularization then transforms the one-hot label vector into a continuous label distribution, which reflects the relations between the target class and other classes based on the constructed class hierarchy. Thus, the model can learn the inter-class relational information, which helps enhance the learning of both old and new classes. Experimental results on both public datasets and a real-world fault diagnosis dataset verify the effectiveness of the proposed method. 
\end{abstract}

}

\maketitle

\IEEEdisplaynontitleabstractindextext

%
\IEEEpeerreviewmaketitle

\IEEEraisesectionheading{\section{Introduction}\label{sec:introduction}}

%
%
%
%
\IEEEPARstart{T}{raditional}
deep neural networks (DNNs) are generally applied in closed scenes, where the entire data to be learned are presumed to be fully obtained during training~\cite{Wang2021review, Li2021AdaptiveNS}. In real-world applications, new classes often appear over time, and the DNNs should be capable of adapting their behaviors dynamically without having to be re-trained from scratch. For instance, autonomous vehicles are required to quickly identify new road signs when entering a new country; fault diagnosis systems should learn to recognize new fault types that occur owing to equipment aging issues~\cite{wang2021coarse}. To this end, class incremental learning (CIL) is proposed and has been attracted much attention in recent years. DNNs need to dynamically update on the new data instead of being trained on a collective dataset with all classes. 
An intuitive approach to CIL is to fine-tune the DNN model for old classes on new data~\cite{li2018learning}. Unfortunately, this leads to catastrophic forgetting, where the DNNs significantly lose their ability to identify the old classes~\cite{wu2019large, hou2019learning}. 

\begin{figure}[!ht]
\centering
\includegraphics[width=0.48\textwidth]{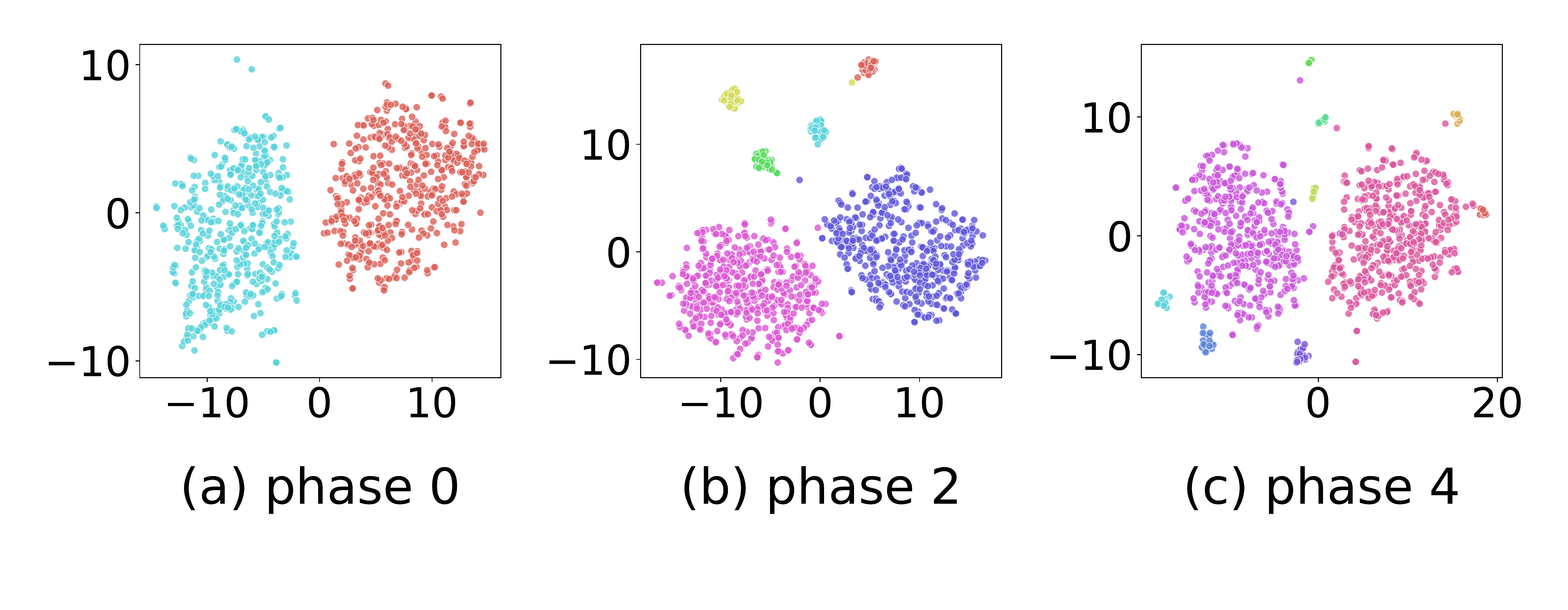}
\includegraphics[width=0.48\textwidth]{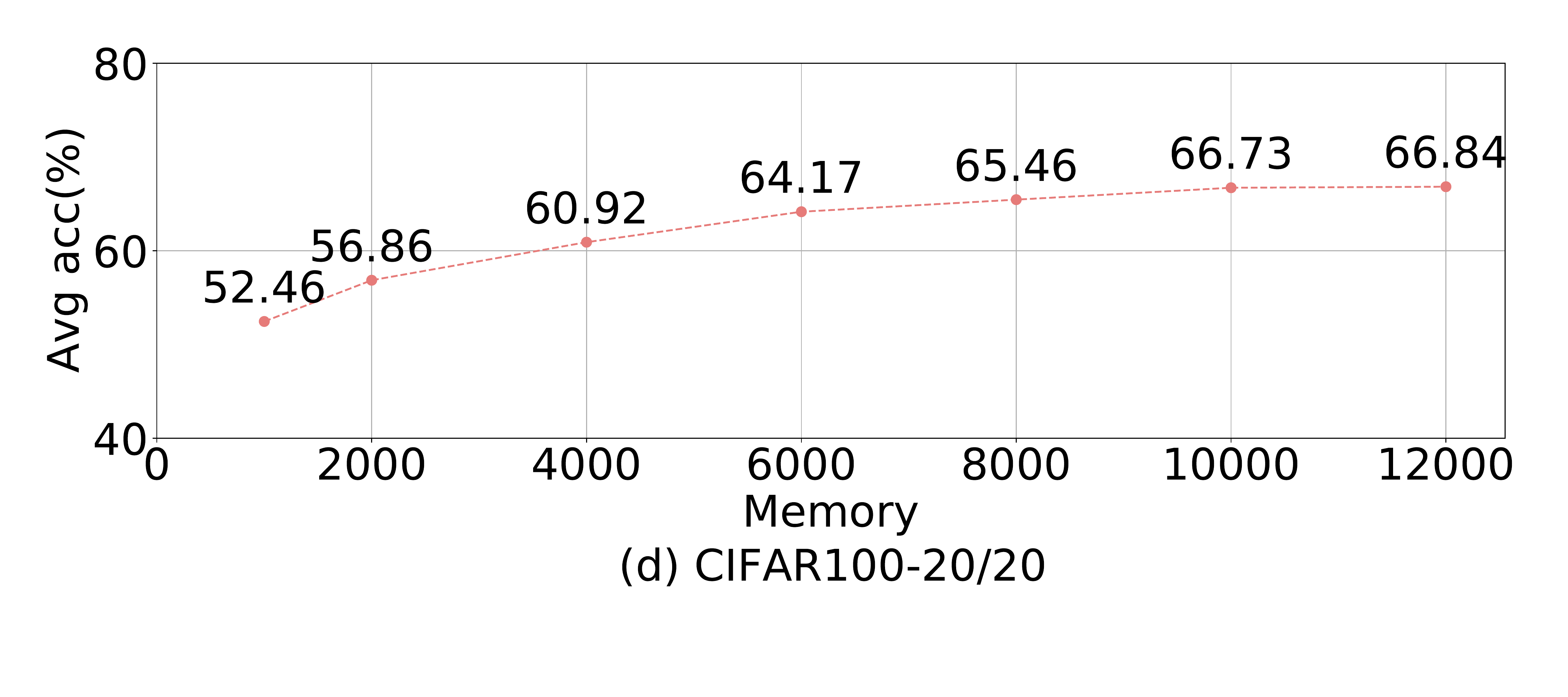}
\caption{Subfigures (a-c) visualize the changing distribution of samples via t-SNE~\cite{maaten2008visualizing} with CIFAR10-2/2 setting, where two classes of samples are presented initially while another two classes are given in each incremental phase. Phase 0 is the initial phase, and phase n is the n-th incremental phase. Note that the memory for old samples is fixed in each incremental phase. Hence data imbalance between old and new samples becomes severe with additional incremental phases. 
Subfigure (d) shows the average accuracy of CIFAR100-20/20 for the baseline method under various memory settings. It shows that large memory for old samples would help improve the performance, implying the data imbalance problem is adverse to incremental learning.
}
\label{fig_datadistribution}
\end{figure}

Significant efforts have been made to address this problem~\cite{hou2019learning,douillard2020podnet,yan2021der}. Typically a few exemplars of old classes are reserved for making the model retain previous knowledge~\cite{yan2021der,chaudhry2019using}. Although CIL tasks have significantly improved with such an approach, a major problem remains: \emph{data are heavily imbalanced between old and new classes owing to the memory limitation of old data.} As displayed in Fig. \ref{fig_datadistribution} (a-c), data imbalance is further exacerbated by more incremental iterations. This imbalance limits the performance of the model. 
Fig. \ref{fig_datadistribution} (d) empirically shows that the performance improves as the memory of old data augments.

Several methods were proposed to address the problem of data imbalance. Since the model would be strongly biased towards new classes due to its overwhelming number of samples, different strategies were designed to rectify the bias. Typically, Wu et al. proposed a BiC approach that adds a linear layer after the final fully connected layer for bias correction~\cite{wu2019large}. Hou et al. designed a LUCIR approach that enforces balanced magnitudes across all classes by using a cosine normalization component~\cite{hou2019learning}. Although these methods show great improvement in CIL, they rely heavily on the assumption of the bias relation between old and new classes. Specifically, BiC assumes a linear relation of the bias, whereas LUCIR requires that the feature vectors of old and new classes are aligned within a sphere. In real-world applications, such assumptions can hardly hold and would thus harm the performance of the model. 

In this paper, we propose a novel method, Multi-Granularity Regularized re-Balancing (MGRB), to address the problem of data imbalance based on a common knowledge distillation baseline. Firstly, re-balancing strategies have been widely used to handle data imbalance without assumptions on data relation. Hence, such a strategy is employed to deal with the imbalance problem and prevent the assumptions of the bias relation between old and new classes. However, we empirically discovered that although the performance of old classes can be improved, the model would under-fit new data, consequently leading to performance degradation. To this end, a multi-granularity regularization term is designed to consider the correlations of all classes, thereby boosting the performance of both old and new classes.
In real-world applications, the class labels are often organized into a tree structure, where coarse-grained nodes represent more general concepts than fine-grained nodes. The ``granularity" refers to the information organized at a certain level in terms of semantics or visual features, reflecting the type of relations between different classes.
Specifically, at the beginning of each incremental learning phase, a class hierarchy is first constructed by grouping semantically or visually similar classes. Subsequently, the correlations of other classes with respect to a specific class are computed according to the similarity within the hierarchy. Such correlations are encoded in the label information by converting the conventional one-hot label vector into a continuous label distribution. The model learns the multi-granularity correlations through gradient update, which boosts the learning of both old and new classes.

To verify the effectiveness of the proposed method, we conduct experiments on multiple public datasets as well as a real-world fault diagnosis application, the fault diagnosis task. Various incremental learning settings are performed, and the results show the superior performance of the proposed method.
The main contributions of this study are summarized as follows:
\begin{itemize}
    \item We propose a novel method, multi-granularity regularized re-balancing, that is assumption-agnostic for the bias relation between old and new classes.
    \item We empirically find that the vanilla re-balancing strategies under-fit new classes despite facilitating retaining the previous knowledge. Therefore, a multi-granularity regularization term is designed by constructing the class hierarchy and embedding the correlations of classes based on the hierarchy in the learning process.
    \item Experimental results show that the proposed method can achieve state-of-the-art performance in CIL tasks and consistently improve the state-of-the-art methods without increasing the complexity of the models. Typically, the proposed method can obtain about 2.57$\%$ and 9.99$\%$ improvement in average accuracy over the comparison methods on public and real-world scenes, respectively.
\end{itemize}

We first introduce related studies in Section \ref{sec:Related Work} and present our method in Section \ref{sec:Method}. In Section \ref{sec:Experiments}, we report and analyze experiments on public datasets and fault diagnosis dataset collected for real-world applications. Finally, we summarize the study in Section \ref{sec:Conclusion}. Code is available at 
\href{https://github.com/lilyht/CIL-MGRB}{https://github.com/lilyht/CIL-MGRB}.

\section{Related Work}
\label{sec:Related Work}

This work is relevant to class incremental learning, re-balancing, and hierarchical modeling methods. In this section, we introduce these related studies.

\begin{figure*}[htb]
    \centering
    \includegraphics[width=\textwidth]{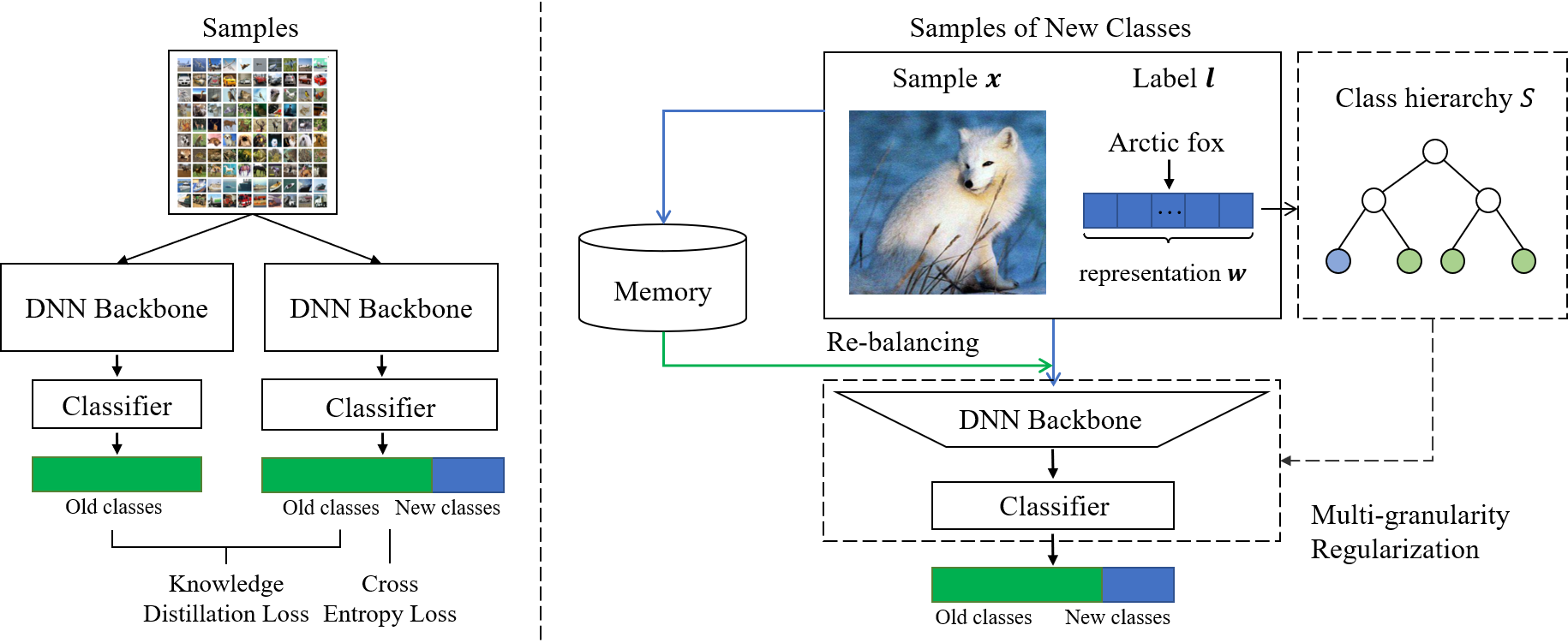}
    \caption{The figure on the left shows the structure of the baseline method, and the figure on the right is the overview of our MGRB method. The proposed model has two significant parts compared with the baseline: (1) re-balancing modeling. We use the re-balancing strategies during training to alleviate the influence of data imbalance; and (2) multi-granularity regularization. Because only using re-balancing strategies would lead to under-fitting of the new classes. Thus, a multi-granularity regularization term is designed to make the model consider class correlations. The class hierarchy is first obtained by using the ontologies of datasets or clustering the semantic word vectors. Subsequently, the one-hot label is converted to a continuous label distribution and optimized using Kullback-Leibler divergence loss. Through end-to-end learning, both old and new classes can be better learned.
    }
    \label{fig_mode}
\end{figure*}

\subsection{Class Incremental Learning Methods}

\textbf{Rebalancing-based Approaches} are the most relevant type of studies to our work in CIL. 
Because the number of samples of new classes is significantly larger than that reserved for old classes, the model is mostly heavily biased towards new classes, leading to the loss of previous knowledge. To this end, many researchers have attempted to rectify this bias during training. Wu et al.~\cite{wu2019large} added a linear layer after the fully connected layer and trained it using a balanced validation set to rectify the biases generated from the previous imbalanced training data. Hu et al.~\cite{hu2021continual} and Yan et al.~\cite{yan2021der} proposed correcting bias by adjusting the network modules. They provided a unique head for each class and added a new feature extractor at each incremental step, respectively. 
Hou et al.~\cite{hou2019learning} proposed replacing the original softmax layer with a cosine normalization layer. This ensured the prediction of new classes and old classes with the same amplitude.
Although these methods obtained significant performances, a major problem remains: the bias relation between old and new classes is pre-defined. This may not hold in complex real-world applications and thus cannot be effectively applied in such scenarios.

\textbf{Rehearsal-based Approaches} store a small number of samples from previous tasks by applying a sampling strategy or data augmentation. The use of exemplar rehearsal for CIL was first proposed by Rebuffi et al.~\cite{rebuffi2017icarl}. They proposed a herding strategy that selects the top m samples closest to the average vector of the class and adds them to the sample set. Liu et al.~\cite{liu2020mnemonics} and Wu et al.~\cite{wu2021curriculum} also applied herding to their work. Kang et al.~\cite{kang2020decoupling} further proposed several sampling criteria in their work. Liu et al.~\cite{liu2020mnemonics} selected and saved samples at the edge of the sample space to save. Chaudhry et al.~\cite{chaudhry2019using} proposed a method for selecting the anchor with a more significant loss and closer to the center of each class. 
The proposed method eliminates the expensive computational cost by randomly selecting exemplars and focusing on resolving catastrophic forgetting by regularizing the model.

\textbf{Regularization-based Approaches} add additional constraints to solving catastrophic forgetting problems. Knowledge distillation~\cite{hinton2015distilling} belongs to a critical regularization method, which transfers knowledge from an old model to a new model by preserving network outputs. Wu et al.~\cite{wu2019large} and Castro et al.~\cite{castro2018end} used knowledge distillation in their models. In addition, other studies also focused on limiting the changes in important weights. Kirkpatrick et al.~\cite{kirkpatrick2017overcoming} controlled the essential parameters that did not deviate too far from the previously trained values. Hou et al.~\cite{hou2019learning} regularized the L2-normalized logits of the previous and current networks based on the cosine similarity to eliminate the bias. They also proposed a less-forget constraint to retain the features of the old samples and the parameters of the classifier by limiting the angle of the features of the sample on the old and new models. Verma et al.~\cite{verma2021efficient} applied a Kullback-Leibler divergence loss function to expand the decision boundaries between the features in the data of the given task. 
In this work, the inter-class information is used to constrain the update of model parameters to ensure that the model can grasp the multi-granularity correlations of classes to learn old and new classes effectively. 

\subsection{Rebalancing Strategies for Imbalanced Modeling}
Re-weighting and re-sampling are two common strategies in imbalanced modeling. In this study, they are used to address the imbalance between old and new data to ensure that the assumption of bias relation can be removed. 

\textbf{Re-weighting Strategies} apply different weight parameters to balance the intensity of sample supervision. Cui et al.~\cite{cui2019class} considered the effective number of samples and designed a class-balanced loss based on an effective number of samples. Lin et al.~\cite{lin2017focal} added a factor of cross-entropy loss to focus training on difficult samples. Tan et al.~\cite{tan2020equalization} proposed a softmax equalization loss.
The suppression of tail classes is ignored when the frequency of rare classes is lower than the threshold.
Castro et al.~\cite{castro2018end} proposed a re-weighting method, called the Dynamic Weight Average method, that averages the weights for each task by considering the rate of change of loss for each task.

\textbf{Re-sampling Strategies} use sampling strategies to expand the number of samples in the sparse class. The oversampling method repeatedly adds the original sample to the sparse sample category. Because it is rare and prone to over-fitting, Castro et al.~\cite{castro2018end} and Zhu et al.~\cite{zhu2021prototype} used data enhancement methods such as rotation, flip, cropping, and mixup~\cite{zhang2017mixup} to increase the diversity of the added samples, while He et al.~\cite{he2008adaptive} and Chawla et al.~\cite{chawla2002smote} proposed achieving sample balance by artificially synthesizing data. Prabhu et al.~\cite{prabhu2020gdumb} proposed a sampler to select the most significant classes and discard random samples while adding samples of new classes. The sampler controls the memory to make the samples of old and new classes more balanced.

Although re-balancing methods are simple and effective, they have been proved to under-fit the majority classes and over-fit the minority classes~\cite{zhou2020bbn,chen2021imagine}.
In this paper, we attempted to use re-balancing methods that are designed to handle data imbalance on the class incremental learning task and discussed whether they could work. We empirically found that simply using such methods significantly decreases performance in new classes. This may result from over-learning old classes and under-fitting new classes due to the re-balancing objective. To this end, this study proposes a regularization term to consider multi-granularity relations so that the performance of both old and new classes can be enhanced via embedding such relations during training.
\subsection{Hierarchical Modeling}

Several studies have shown that in large-scale tasks, categories are often formed as a hierarchical structure, similar to the human brain's hierarchical and abstract memory of different objects~\cite{Zhao2020Recursive}. Hierarchy reflects the refinement of knowledge in the learning process.
Multi-granularity modeling involves structure construction and constructed hierarchy utilization~\cite{zheng2017hierarchical,hu2018review}.

For constructing structures, some build structures based on category confusion, while others build structures according to category similarity. Bengio et al.~\cite{bengio2010label} built a tree structure determined by spectral clustering of the confusion matrix. Whereas Lei et al.~\cite{lei2014learning} constructed a visual concept network and grouped visually similar classes, Zhou et al.~\cite{zhou2013jointly} used spectral clustering to ensure that the same group had strong visual correlations. Apart from clustering techniques, Ray et al.~\cite{ray2010semantic} proposed a dynamic and extendable hierarchy construction method based on semantic information in WordNet and Wikipedia.
Classification with a hierarchy is another problem considered by many scholars. Bertinetto et al. \cite{bertinetto2020making} proposed a hierarchical architecture to make better mistakes in image classification. Deng et al.~\cite{deng2012hedging} and Wang et al.~\cite{9380703} directly used the hierarchy to implement tasks from top to bottom. The entire task needed to be divided from top to bottom based on the hierarchy to solve the complex original problem. 
Some studies transformed hierarchy information into labels. Bertinetto et al.~\cite{bertinetto2020making} proposed a label-embedding approach and a hierarchical cross-entropy loss, replacing the one hot label with a soft label that contained hierarchy information. 
Some scholars applied multi-granularity modeling to human parsing. Wang et al.~\cite{wang2021hierarchical} modeled the human parsing with a hierarchical structure since the human body presents a highly structured hierarchy. Zhou et al.~\cite{zhou2021differentiable} considered the instance-aware person part parsing as a multi-granularity person representation learning task.

In this study, we use the ontology, semantic information, or visual feature to construct the class hierarchy to obtain the multi-granularity correlations of classes. The correlations are subsequently transformed into a label distribution that serves as the optimization target. Through gradient updates, these correlations are embedded in the model to boost the learning of both old and new classes.

\section{Method}
\label{sec:Method}

As illustrated by Wu et al.~\cite{wu2019large} and Hou et al.~\cite{hou2019learning}, the weights and biases of the classifier for new classes are significantly higher than those of old classes; thus, they mistakenly classified samples of old classes as new classes. 
However, the previous bias correction work was restricted to limited situations because they had an assumption of biases. In this study, we are committed to making our method available for real-world applications. 

We propose to solve the catastrophic forgetting of incremental learning by a multi-granularity regularized re-balancing method. The framework of the proposed model is shown in Fig. \ref{fig_mode}. The re-balancing modeling alleviates the influence of data imbalance. However, it is empirically found that it would result in the under-fitting of new classes. To address such a problem, we use a multi-granularity regularization to enhance the learning of both old and new classes.

\subsection{Baseline Method}
\label{sec:Method_Baseline}
Masana et al.~\cite{masana2020class} defined an incremental learning problem. We denote the dataset $D=\{D_1, D_2, ...,D_C\}$, where $D_i = \{(x_1,y_1), (x_2, y_2), ..., (x_n, y_n)\}$, represents the sample set of class $i$. The baseline method uses cross-entropy loss to penalize the classification error:

\begin{equation}\label{eq_ce_1}
L_{ce}=-\sum_{D} \sum_{i}^{C} \log \left(\frac{\exp \left(z_{i}\right)}{\sum_{j=1}^{N} \exp \left(z_{j}\right)}\right).
\end{equation}

The knowledge distillation method is also applied to the baseline method. Specifically, the model in the previous round is used as a teacher model, and the model that needs to learn in this round is called the student model. 
The input of the student model is $D_{t}=D_{\text {old }}^{\prime} \cup D_{\text {new }}$, where $D_{\text {old }}^{\prime}$ represents the saved exemplars. The corresponding output is $z=\{z_{1}, z_{2}, \ldots, z_{N_{\text {old }}+N_{\text {new }}}\}$. The old data pass through the teacher model, and the output $z^{\prime}=\{z_{1}^{\prime}, z_{2}^{\prime}, \ldots, z_{N_{\text {old }}}^{\prime}\}$. Before passing through the softmax layer, each item of the probability in the output of the student model and teacher model is divided by the temperature $T$, and record the results as the soft label $\pi(z)$ and $\pi(z^{\prime})$, respectively.
Subsequently, the output $\pi(z)$ is cut to obtain the predicted probability of the former $N_{old}$ classes, and the knowledge distillation loss is calculated. The knowledge distillation loss function is defined as follows:

\begin{equation}\label{eq_LD}
L_{D}=-\sum_{i=1}^{N_{old}} \pi\left(z_{i}^{\prime}\right) \log \left(\pi\left(z_{i}\right)\right), 
\end{equation}
\begin{equation}\label{eq_pizi}
\pi\left(z_{i}^{\prime}\right)=\frac{\exp \left(\frac{z_{i}^{\prime}}{T}\right)}{\sum_{j=1}^{N_{\mathrm{old}}}\left(\frac{z_{j}^{\prime}}{T}\right)}, \\
\pi\left(z_{i}\right)=\frac{\exp \left(\frac{z_{i}}{T}\right)}{\sum_{j=1}^{N_{\text {old }}}\left(\frac{z_{j}}{T}\right)}.
\end{equation}

\subsection{Re-balancing Modeling}

Considering that a severe data imbalance exists in the training set, we propose re-weighting and re-sampling methods to address this problem. 

\subsubsection{Re-weighting Approach}
\label{sec:method_CB}

We utilize a class-balanced classification loss function proposed by Cui et al.~\cite{cui2019class}. They allocated different weights to each category according to the effective number of samples.
Intuitively, this re-weighting method is also suitable for CIL. After the arrival of new classes, the proportion of old classes becomes smaller, transforming into relatively sample-sparse classes. Therefore, the re-weighting method can strengthen the supervision of old classes and adjust the supervision intensity of the old classes to the same level as new classes.
The class-balanced loss is defined as: 
\begin{equation}\label{eq_cbloss}
L_{CB}(z,y) = -\sum_{i=1}^N  \frac{1-\gamma}{1-\gamma^{n_i}} log(\frac{exp(z_i)}{\sum_{j=1}^N exp(z_j)}),
\end{equation}
\begin{equation}\label{eq_gamma}
\gamma = \frac{N-1}{N},
\end{equation}

where $z_i$ is the logits of the network, $N$ represents the total number of classes, and $n_i$ is the sample number of each class.

\begin{figure*}
    \centering
    \includegraphics[width=\textwidth]{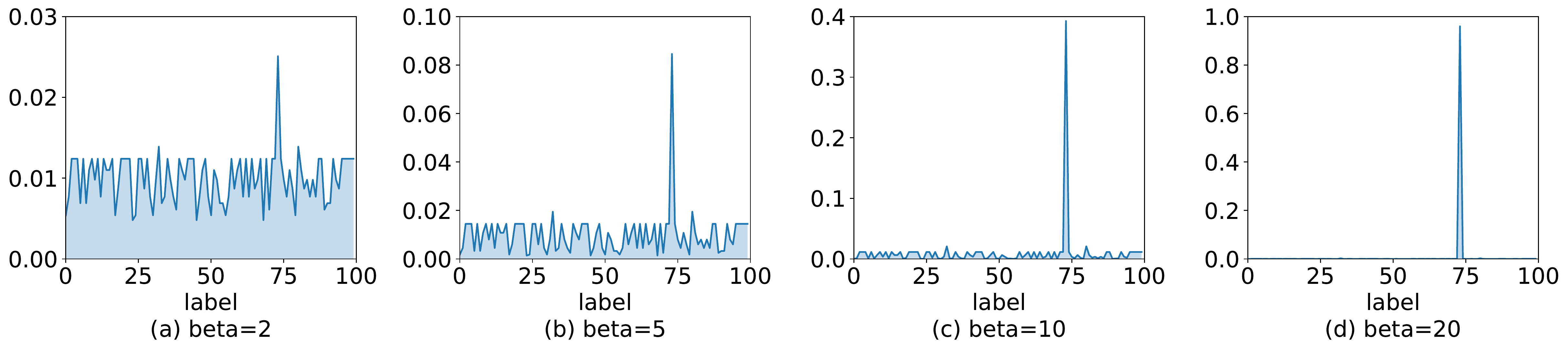}
    \caption{The influence of parameters $\beta$ on the label distribution (the ground truth $g$ is ``fence", and the index is 73) with the miniImageNet. We can observe that index 73 has the highest value. The secondary highest indexes are index 32 (bookstore) and index 80 (tobacco shop). These three classes are under the subtree ``entity - physical entity - object - whole - artifact - structure" of the ontology. It can also be found that as the parameter $\beta$ increases, the distribution of labels becomes sharper.}
    \label{fig_labeldistribution}
\end{figure*}

\subsubsection{Re-sampling Approach}
\label{sec:method_DL}

We also address the imbalance in the re-sampling aspect. We generate a balanced exemplar set by randomly selecting from the memory and newly obtained data. The balanced exemplar set contains two parts: exemplars for old classes and exemplars for new classes. Each part is a subset of the saved samples or newly obtained. Note that the balanced example set and the training set do not overlap and are divided before training. Specifically, the latter part consists of the same number of samples per class as the former.

Decoupling training separates a module from the whole network. The research of Kang et al.~\cite{kang2020decoupling} showed that decoupling learning could obtain higher accuracy in unbalanced situations. Subsequently, we apply a decoupling training strategy using the balanced exemplar subset.
We decouple the feature extraction layer from the classifier after training with imbalanced samples, freeze the parameters of the feature extractor, and use the re-sampled balanced subset to retrain the classifier.
This decoupling strategy is applied in all phases except the initial phase because the samples in the initial phase are balanced.

\subsection{Multi-granularity Regularization}
\label{sec:Method_Muli-granularity}
Existing methods use a one-hot label vector for each sample, in which the ground truth class $g$ is set to one and the others are set to zero. Optimizing such a vector enables the model to treat classes other than the ground truth class equally. In this case, re-balancing strategies reduce the update of new classes to alleviate the degree of data imbalance. However, this results in the under-fitting of new classes, which is empirically proven in Section \ref{sec:Experiments}. One possible way to solve this problem is to enhance learning new classes when updating samples of old classes. To this end, inspired by Bertinetto et al. \cite{bertinetto2020making}, we use a multi-granularity regularization term that converts the one-hot label vector into a continuous soft label distribution and can consider the correlations between old and new classes during incremental learning. By optimizing the prediction of the model to approximate the constructed label distribution, the learning of old and new classes can be boosted simultaneously.

\subsubsection{Extraction of Hierarchical Structure}
\label{sec:Method_extract}

A hierarchical structure is required to obtain the multi-granularity correlations of classes. According to the specific situation of datasets, the approaches of constructing hierarchy \textbf{$S$} can be divided into three types: ontology, semantic hierarchical clustering, and visual hierarchical clustering. If the dataset has its coarse-grained information, the hierarchy is directly extracted. If not, the hierarchies need to be learned through semantic or visual clustering. Note that we only use the information from the current phase.

\textbf{Ontology.} This method is suitable for multi-granularity datasets or datasets with knowledge structures. 
Samples with both fine-grained and coarse-grained labels can be considered a multi-granularity dataset, which can be organized into a multi-layer hierarchy with coarse-grained nodes at the top and fine-grained nodes at the bottom. This structure indicates ownership. For example, in CIFAR100, ``bottles" and ``can" belong to the same coarse-grained node, and ``bottles'' and ``bridges'' belong to different coarse-grained nodes. In this way, the classes under the same coarse-grained node indicate a closer relationship, while the two classes belonging to different coarse-grained nodes have further inter-class distance. The ontology is updated by adding new branches when future classes arrive.

\begin{figure}[htbp]
    \centering
    \includegraphics[width=0.48\textwidth]{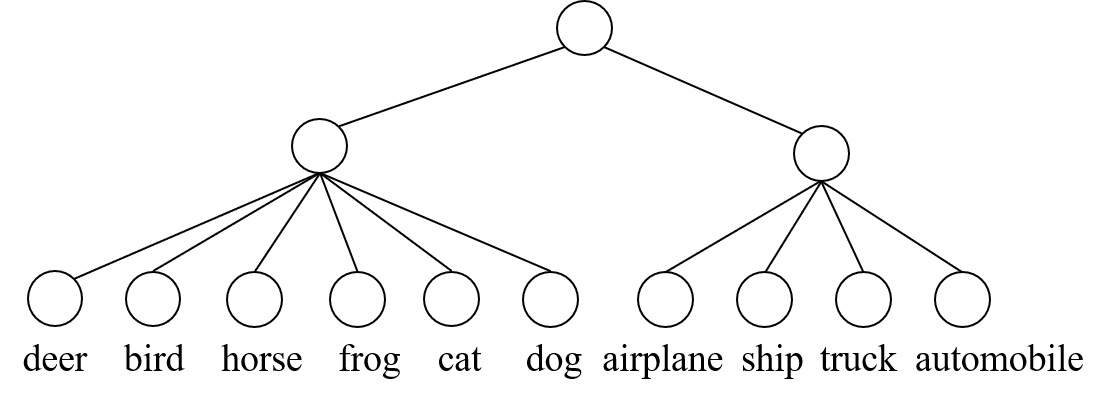}
    \caption{The hierarchy of CIFAR10 is generated by semantic hierarchical clustering when the number of clusters is set to two. It can be observed that ten classes are grouped according to semantic similarity. The coarse-grained nodes of the two clusters can be considered to represent animals and vehicles, respectively. Fine-grained classes in the same cluster are semantically similar.}
    \label{fig_cifarhier}
\end{figure}

\textbf{Semantic Hierarchical Clustering.} For flat datasets that do not provide a hierarchy, such as CIFAR10 or some self-built datasets, using semantic clustering to build a hierarchy is also effective.
GloVe (Global Vectors for Word Representation), proposed by Pennington et al.~\cite{pennington2014glove}, is a word embedding method based on global word frequency statistics. It is implemented by mapping words to a high-dimensional space. A pre-trained word vector library provided by Google is available on the website\footnote{https://code.google.com/archive/p/word2vec/}, and each word is embedded into a 300-dimension feature space. We denote the word vector of the sentiment label $\boldsymbol{l}$ as $\boldsymbol {w}$.
After obtaining the representation of the words in the feature space, the K-means clustering algorithm~\cite{krishna1999genetic} is utilized to simply cluster the semantic vectors $\boldsymbol {W} = \{\boldsymbol {w_1}, \boldsymbol {w_2}, ..., \boldsymbol {w_n}\}$ to $k$ coarse-grained groups. The optimization formula for the K-means algorithm is as follows:

\begin{equation}\label{eq_kmeans}
\underset{S}{\operatorname{argmin}} \sum_{i=1}^{k} \sum_{\boldsymbol{w} \in U_i}\|\boldsymbol{w}-\boldsymbol{\mu_i}\|^{2},
\end{equation}
where $k$ is the number of clusters. $\boldsymbol {w}$  represents the semantic vector of labels, and $\boldsymbol{\mu_{i}}$ represents the mean of the word vectors of cluster $U_i$.

Based on this, a two-level hierarchy $S$ is constructed. Note that the hierarchy can be more than two levels. However, this study uses a two-level hierarchy for simplicity. The generated hierarchy of CIFAR-10, when grouped into two clusters, is shown in Fig. \ref{fig_cifarhier}.

Note that when using semantic hierarchical clustering, only the information of the appeared classes can be used. In other words, the hierarchy is dynamically built and completed during incremental learning phases.

\textbf{Visual Hierarchical Clustering.} Semantic clustering inevitably requires additional information, for example, a word vector library. Therefore, for more general cases, we further propose a multi-granularity hierarchical construction method using visual feature clustering. Because we have a well-trained model that handles old classes, it is possible to use this model to extract visual features for clustering. Specifically, before each new incremental phase starts, the previous feature extractor $\Phi$ is used to obtain the features of all samples in the memory and the newly arriving data. 
\begin{equation}\label{feature_extract}
\boldsymbol{v_i} = \frac{1}{p} \sum_{j=1}^{p} \Phi(\boldsymbol{x_j}),
\end{equation}
where $\boldsymbol{v_i}$ represents the average feature for class $i$; $p$ is the number of exemplars per class, and $\boldsymbol{x_i}$ is the input data. The average of the sample features of each class is considered the representative visual feature of the class. Then, the K-Means clustering algorithm is used to cluster the visual vectors $\boldsymbol {V} = \{\boldsymbol {v_1}, \boldsymbol {v_2}, ..., \boldsymbol {v_n}\}$ to $k$ groups. Similarly, the optimization formula for the K-means algorithm can be written as:

\begin{equation}\label{eq_visual_kmeans}
\underset{S}{\operatorname{argmin}} \sum_{i=1}^{k} \sum_{\boldsymbol{v} \in U_i}\|\boldsymbol{v}-\boldsymbol{\mu}_i\|^{2},
\end{equation}
where $k$ is the number of clusters and $\boldsymbol{\mu}_{i}$ is the mean of the visual feature vectors of cluster $U_i$. In this way, we can obtain coarse-grained concepts and form a multi-granularity structure.

\subsubsection{Multi-granularity Regularized Loss Term}
\label{sec:Method_mgloss}

With the extracted hierarchy, the original one-hot labels are converted to soft labels to reflect the correlations between categories. The label containing hierarchical information is written as:

\begin{equation}\label{eq_hierlabel}
y_A^M = \frac{exp(-\beta d(A,C))}{\sum_{B\in N} exp(-\beta d(B,C)},
\end{equation}

\begin{equation}\label{eq_hierdistance}
d(A, C) = \frac{Height(LCS(A,C))}{max_{B \in N} Height(B)}, 
\end{equation}
where $C$ represents the node of the real label, $A$ represents the current class node, $M$ stands for the multi-granularity label, and $N$ represents the set of leaf nodes. $\beta$ is a hyperparameter of the function, with a value range of (0, $+\infty$), which controls the distribution of labels. The formula $d(N_i,N_j)$ measures the distance between two fine-grained classes in the hierarchy. In Equation \eqref{eq_hierdistance}, LCS (lowest common subtree) represents the smallest common subtree of two nodes. $Height(B)$ represents the height of a subtree whose root is node $B$. The hierarchical label is noted as $Y = \{y_{N_1}^M, y_{N_2}^M, ..., y_{N_n}^M\}$.

The above formula has two properties.
First, $\beta$ reflects the intensity of the oversight of hierarchical information. The smaller the value of $\beta$ becomes, the more evenly distributed hierarchical structure tags. If $\beta$ increases, the label distribution becomes sharper.
Second, classes with stronger correlations obtain higher responses. The larger the differences between the current node and the target label node in the branch, the less their intra-class relationship is, then $d(A,C)$ becomes larger, resulting in a smaller value of $y_A^M$. Nodes having more significant correlations with the target class can obtain higher corresponding values in the generated label than the nodes with fewer correlations.

Fig. \ref{fig_labeldistribution} visualizes the label distribution on miniImageNet, and shows the influence of parameter $\beta$ on the label distribution. Note that although the label distribution is extremely sharp in Fig. \ref{fig_labeldistribution} (d), the soft label is identical to the one-hot label. We use the Kullback-Leibler divergence~\cite{kullback1951information} to calculate the differences between $y_A^M$ and prediction probability $q_i$. The multi-granularity loss function is formulated as
\begin{equation}\label{eq_hierloss}
L_H = \sum_i \sum_{j=1}^N Y_j log(\frac{Y_j}{q_j}).
\end{equation}
The overall loss in the proposed model is:
\begin{equation}\label{eq_overall_loss}
L = \lambda L_{CB} + \alpha(1-\lambda) L_D + L_H,
\end{equation}
where the scalar $\lambda$ is usually set to $\frac{N_{old}}{N_{old}+N_{new}}$ to adjust the constraints on previous knowledge. $\alpha$ is used to adjust the scale between classification loss and knowledge distillation loss.

\noindent\textbf{Gradient Analysis}.
\label{sec:Method_derivation}
From the perspective of gradient, we analyze the procedure by which the multi-granularity regularization boosts the learning of old and new classes. In one-hot encoding, only the real class is set to one, and the value of the other classes is zero. The traditional cross-entropy loss is written as:
\begin{equation}\label{eq_ce_2}
L_{ce} = - \sum_i \sum_{j=1}^N p_j log(q_j),
\end{equation} 
where
\begin{equation}\label{eq_pi}
p_j = \begin{cases}
1 & \text{$j==g$}\\
0 & \text{others},
\end{cases}
\end{equation}
and $g$ represents the true label of sample i.
In the proposed multi-granularity loss, the transformed label distribution contains the information on the hierarchical structure (see Equation \ref{eq_hierlabel}).
We use Kullback-Leibler divergence to approximate the multi-granularity label distribution $Y(x)$ and the predicted distribution $\hat{Y}(x)$. The final optimal parameter $\theta$ of the model is:

\begin{equation}\label{eq_theta}
\theta = \mathop{argmax}\limits_{\theta} \sum_i \sum_{A \in N} y_A^M log(\frac{y_A^M}{\hat{y}_A^M}),
\end{equation}
where $\hat{y}$ is the logits after the softmax layer. 

The proposed term treats classes other than the ground-truth classes differently concerning the correlations of classes. Classes closer to the ground-truth classes obtain higher values in the label distribution and, consequently, larger gradient updates. In this manner, the multi-granularity correlations among classes are adequately reserved in our approach, enhancing the learning of both old and new classes.

\subsubsection{Discussions on the Multi-granularity Regularization Term}

(1) Difference in label smoothing. The designed multi-granularity regularization term is similar to label smoothing, and here we discuss the difference between them. Label smoothing encourages samples to be equidistant from other classes, which prevents overconfidence and improves model generalization. However, label smoothing can not capture class relationships because it treats classes except the correct class as the same error. By contrast, the proposed multi-granularity regularization constructs a label distribution according to the hierarchical relationship between classes. Therefore, besides facilitating generalization, the proposed method can use the multi-granularity relations of other classes to help the learning of a specific class, which is beneficial for solving the problems of both data scarcity for old classes and under-fitting the new classes.

(2) Difference in the path-based hierarchy-aware method. Another method of embedding multi-granularity relations is hierarchical cross-entropy loss (HXE) proposed by Bertinetto et al.~\cite{bertinetto2020making}. The HXE loss calculates the conditional probability of each node on the chain to obtain the final probability of the leaf node. The advantage of considering class probabilities is that no direct knowledge of the hierarchy is required during inference. However, the disadvantage is that such a top-down classification would lead to error propagation. i.e., the misclassification on a certain level would pass such an error to the next level. The multi-granularity regularization term encodes the hierarchical information into a soft label. The advantage of the method is that the hierarchical structure information is embedded in the label, which is easy to calculate and apply. The disadvantage is that the detailed path is not concerned. Even if the leaf nodes of the same depth belong to different subcategories, they can be mapped to the same value in the label, thus losing such information during modeling.

\section{Experiments}
\label{sec:Experiments}
We conduct experiments on four commonly used public datasets and a real-world fault diagnosis dataset to verify the effectiveness of the proposed method. In such a scenario, some fault types cannot be collected at the beginning. However, they emerge gradually during the operation of the machine with the iterative update. This comprehensively tests the adaptive industrial task performance of the model.

\subsection{Experiments on Public Datasets}

\subsubsection{Experimental Setups}
\label{sec:Public_Experimental_setup}

We compared the proposed method with several methods on four image datasets: CIFAR10, CIFAR100~\cite{krizhevsky2009learning}, miniImageNet and ImageNet-Subset~\cite{rebuffi2017icarl}. CIFAR10 consists of 10 classes, each containing 5,000 32 $\times$ 32 images for training and 1,000 images for testing. CIFAR100 consists of 100 classes, each containing 500 32 $\times$32 images for training and 100 images for testing. MiniImageNet also consists of 100 classes, each containing 500 84 $\times$ 84 images for training and 100 images for testing. ImageNet-Subset has 100 classes, each containing 1300 images. The 100 classes are selected from ImageNet in the same way as~\cite{rebuffi2017icarl,wu2019large}.
We split the datasets into two different methods. In the first splitting method, datasets are split equally. Whereas in the second splitting method, half of the total classes are divided as the initial phase classes, after which the remaining classes are divided into $t$ incremental batches. This study uses $n/m$ to represent learning n classes in the initial phase and learning m classes in the following phases until all classes are learned.

We used incremental accuracy and average incremental accuracy to evaluate the models. The average incremental accuracy metric is the average accuracy of each phase of incremental learning, including the initial phase. Our models are implemented using PyTorch. All the experiments were run on one NVIDIA RTX3090 GPU with 24 GB. 

To validate the effectiveness of the proposed method, we compared the proposed method with classical and latest methods. 
\textbf{LwF}~\cite{li2018learning} adds a new classifier for every newly coming task. The CNN layer and the parameters of the old task classifiers are frozen, and the classifier of the new tasks is trained until they converge. Then the parameters of the entire network are trained together.
\textbf{iCaRL}~\cite{rebuffi2017icarl} adopts a herding strategy for re-sampling and a nearest-mean-of-exemplars classifier to avoid bias.
\textbf{BiC}~\cite{wu2019large} adds a linear layer to the final fully connected layer for bias correction. It adopts a two-stage training method and uses the balanced validation set to train the linear correction layer in the second stage. 
\textbf{LUCIR}~\cite{hou2019learning} employs cosine normalization, less-forget constraint, and inter-class separation to preserve the previous knowledge and reduce the ambiguities between old and new classes. 
\textbf{Mnemonics}~\cite{liu2020mnemonics} uses model-level and exemplar-level optimizations to optimize model parameters and select examples in turn.
\textbf{PODNet}~\cite{douillard2020podnet} employs an efficient spatial-based distillation loss throughout the model. It also uses a more flexible representation comprising multiple proxy vectors for each class. 
\textbf{AANets}~\cite{liu2021adaptive} learns stability and plasticity using plasticity and stability blocks, respectively, and then gets a final representation by aggregating them.
\textbf{GeoDL}~\cite{simon2021learning} proposes a distillation loss in the geometry aspect. It focuses on the gradual changes of two different tasks and models them with the geodesic flow.
\textbf{DER}~\cite{yan2021der} increases feature dimensions when new classes appear and proposes a channel-level pruning strategy to constrain the parameters. They introduce an auxiliary loss to learn diverse features.
In this paper, the protocols of all comparative experiments are unified. For example, each method is unified to save a fixed number of samples per class, and the learning order are the same.
Codes of the comparison methods in original papers are re-run in our experimental environment.

\subsubsection{Implementation Details}
\label{sec:Public_implementation_detail}

The hierarchy of CIFAR10 was built by semantic clustering and was grouped into two clusters. The hierarchy of semantic clustering was illustrated in Fig. \ref{fig_cifarhier}. For CIFAR100, we used its original structure, which is a two-level hierarchical structure. It contains 20 coarse-grained nodes, and each coarse-grained node consists of five fine-grained categories. For miniImageNet and ImageNet-Subset, we used its WordNet hierarchy. We also provided the results of using visual clustering of the first three datasets.

We employed a ResNet32 architecture (without pre-training on ImageNet)~\cite{he2016deep} for CIFAR10 and CIFAR100. We used a ResNet18 architecture (without pre-training on ImageNet) for miniImageNet and ImageNet-Subset as the backbone network. We adopted an SGD optimizer with a weight decay of 2e-4 and a momentum of 0.9. The number of epochs for the imbalance learning for CIFAR10, CIFAR100, miniImageNet and ImageNet-Subset were set to 200, 200, 100 and 100, respectively. For CIFAR10 and CIFAR100, the batch size was set to 128. The learning rate began at 0.01 and was divided by 10 at 70 and 140 epochs. For miniImageNet and ImageNet-Subset, the batch size was set to 128 and 64, respectively. The learning rate both started from 0.01 and was divided by 0.1 every 30 epochs. 

In this paper, we split CIFAR10 into 1/1, 2/2, and 5/1. The CIFAR100 and miniImageNet were both split into 10/10, 20/20, 50/5, and 50/10. The ImageNet-Subset was split into 50/5 and 50/10. We set $\beta$ to 20 and $\alpha$ to 1.0 for CIFAR10 under all split rules. For CIFAR100, we set $\alpha$/$\beta$ to 0.1/20, 1.0/20, 1.0/50, 1.0/50 in order shown in Table \ref{tab_avgacc_public_cifar100}. For miniImageNet, we set $\alpha$/$\beta$ to 0.1/100, 0.1/100, 0.1/100, 1,0/20  in order. For ImageNet-Subset, we set $\beta$ to 50 and set $\alpha$ to 1.0 under all split rules.
We set the knowledge distillation temperature $T$ to 2, ensuring consistency with the results of other studies. 
In terms of the memory that retains samples of old classes, each class can save 200 samples for CIFAR10 and 20 samples for CIFAR100, miniImageNet, and ImageNet-Subset. We randomly selected samples and saved them as exemplars. The ratio of the training and retraining exemplar set was set to 9:1.

\subsubsection{Results on Public Datasets}
\label{sec:Experiments_evaluation_Public}

\textbf{Evaluation on CIFAR10.}
The experimental results on CIFAR10 are shown in Table \ref{tab_avgacc_public_cifar10}. The Ours-CNN(sem) uses semantic hierarchical clustering, and the Ours-CNN(vis) uses visual hierarchical clustering, with $k$ both setting to 2.
It can be observed that the proposed method outperforms the other methods by 2.86$\%$ and 0.74$\%$ under CIFAR10-2/2 and CIFAR10-5/1 respectively on average accuracy. 
The accuracy of all the methods in the initial phase is close. However, for the average accuracy, the proposed method exceeds BiC by a large margin, which indicates that the knowledge is forgotten more slowly. This is because we considered the bias in the sample distribution and classifier, as well as the correlations of classes. 

\begin{table}[htbp]
\centering
\caption{The last phase accuracy and average accuracy on CIFAR10.}
\label{tab_avgacc_public_cifar10}
\resizebox{0.5\textwidth}{!} {
\begin{tabular}{c|cccccc}
\hline
\multirow{2}{*}{Methods} & \multicolumn{2}{c}{CIFAR10-1/1} & \multicolumn{2}{c}{CIFAR10-2/2} & \multicolumn{2}{c}{CIFAR10-5/1} \\
 & last & Avg acc & last & Avg acc & last & Avg acc \\ \hline
LwF & 47.38 & 57.86 & 52.79 & 64.82 & 59.11 & 69.48 \\
Mnemonics & 74.96 & $\bm{78.90}$ & 80.11 & 84.82 & 83.85 & 86.76 \\
BiC & 68.62 & 67.93 & 71.61 & 74.07 & 72.50 & 76.05 \\
Ours-CNN(sem) & 69.57 & 73.01 & 82.87 & $\bm{87.68}$ & 83.33 & $\bm{87.50}$ \\
Ours-CNN(vis) & 66.90 & 68.60 & 83.10 & 86.31 & 82.45 & 86.83 \\ \hline
PODNet & 29.20 & 42.70 & 53.90 & 68.74 & 58.50 & 74.15 \\
Ours-PODNet & 30.60 & $\bm{44.14}$ & 56.40 & $\bm{69.90}$ & 63.80 & $\bm{75.83}$ \\ \hline
AANets-iCaRL & 82.18 & 82.32 & 85.90 & 84.45 & 85.40 & 84.80 \\
Ours-AANets & 82.59 & $\bm{83.24}$ & 83.78 & $\bm{86.31}$ & 86.06 & $\bm{87.79}$ \\ \hline
DER & 65.98 & 68.01 & 75.52 & 81.83 & 77.34 & 81.77 \\
Ours-DER & 66.32 & $\bm{68.73}$ & 76.41 & $\bm{82.12}$ & 78.12 & $\bm{82.29}$ \\ \hline
\end{tabular}
}
\end{table}

\begin{figure}[htb]
    \centering
    \includegraphics[width=3.4in]{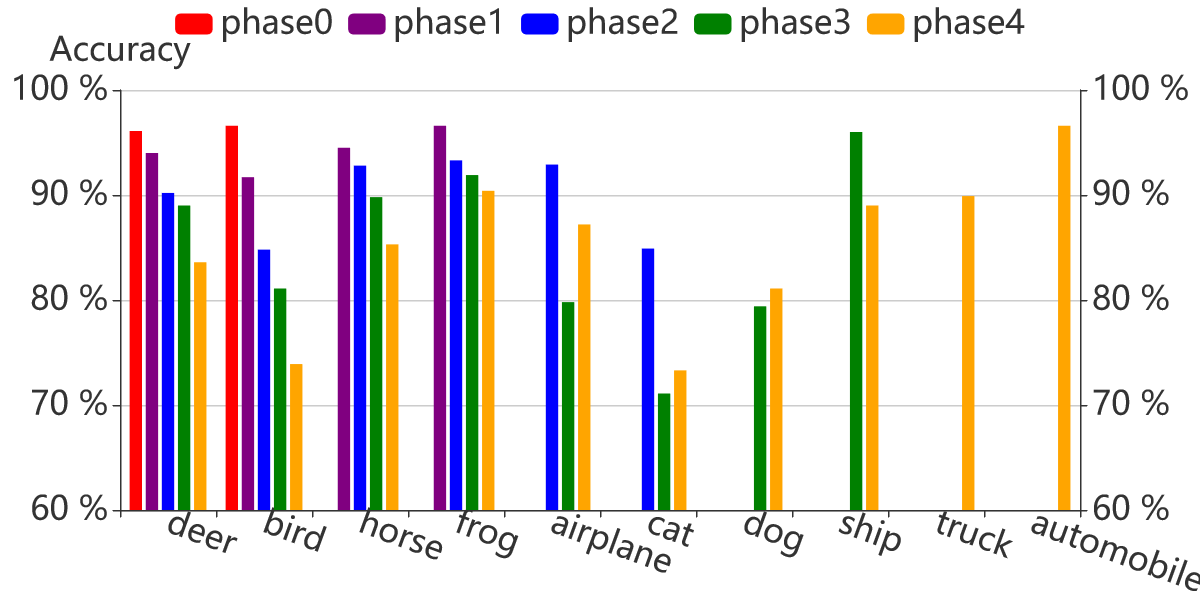}
    \caption{Accuracy of each class of CIFAR10 under the setting of 2/2. The x-axis from left to right indicates the learning order of the classes. For example, the deer and bird first appear in the initial phase and are learned in all phases. The airplane and cat appear in phase 2, and they are learned in phase 2, phase 3, and phase 4.}
    \label{fig_cifar10_acc_each_class}
\end{figure}

Owing to an observed upward trend in the last phase of CIFAR10-2/2, the accuracy of each class in every learning phase was analyzed (see Fig. \ref{fig_cifar10_acc_each_class}). The learning order was generated by setting the seed to 1993~\cite{rebuffi2017icarl}. Generally, almost all classes except for ``airplane", ``cat", and ``dog" present the forgetting tendency when the learning phase is in progress. For the ``airplane" class, the accuracy is 92.9$\%$, 79.8$\%$, and 87.2$\%$ in phase 2, phase 3, and final phase, respectively. The ``cat" and ``dog" classes show similar ascending trends. This phenomenon may be explained by the fact that the first learning of ``dog" in phase 3 confuses the learning of ``cat" learned before. Therefore, owing to limited samples, the knowledge of class ``cat" was forgotten, thus leading to its accuracy degradation. In contrast, the learning of class ``truck" and ``automobile" helps strengthen the knowledge of class ``airplane", resulting in an upward trend as shown in Fig. \ref{fig_cifar10_acc_each_class}.

\begin{figure}[htb]
    \centering
    \includegraphics[width=3.4in]{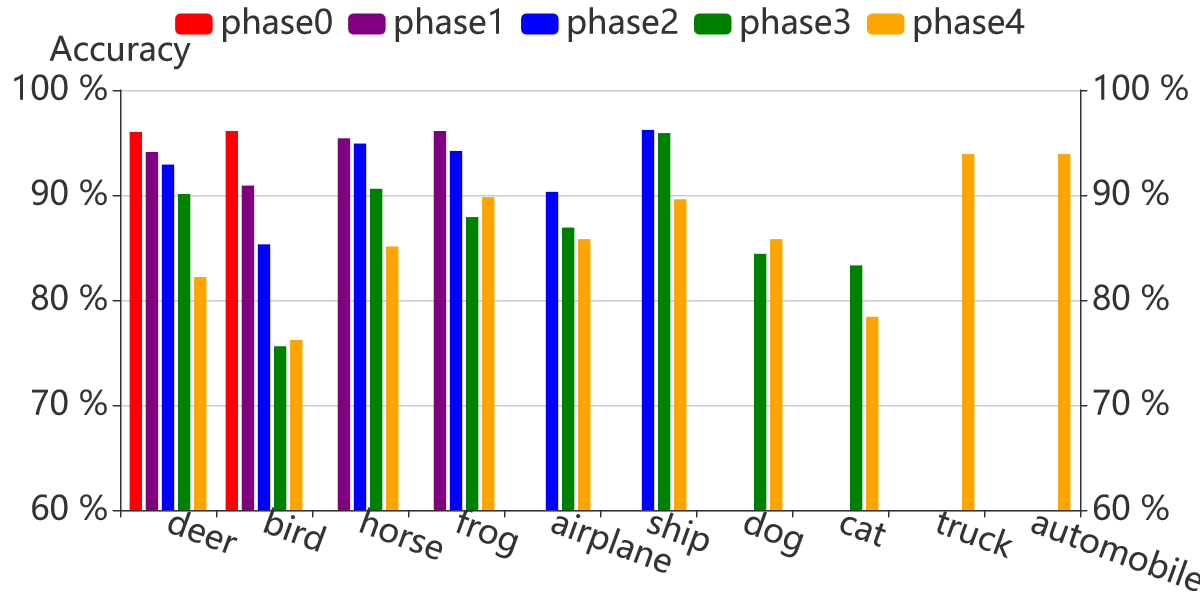}
    \caption{Accuracy of each class of CIFAR10 under the setting of 2/2. The x-axis from left to right indicates the learning order of the classes. Note that the learning order of ``cat" and ``ship" has changed in this experiment.}
    \label{fig_cifar10_acc_each_class-2}
\end{figure}

Furthermore, we performed another experiment that reverses the order of class ``cat" and ``ship". For CIFAR10-2/2, the learning groups change to "airplane" and ``ship" in phase 2, and ``dog" and ``cat" in phase 3. In this manner, the classes of newly learned groups gain closer relationships than before. It can be clearly seen that learning similar classes assists memorization of original knowledge (compare the changes of the accuracy in phase 2 and phase 3 of class ``airplane" and ``ship" in Fig. \ref{fig_cifar10_acc_each_class} and Fig. \ref{fig_cifar10_acc_each_class-2}). When the newly coming classes have a large gap with the previously learned classes in the feature space, the previously learned classes are not easily forgotten (see the changes of the first four classes accuracy in phase 1 and phase 2 in Fig. \ref{fig_cifar10_acc_each_class-2}).

\textbf{Evaluation on CIFAR100, miniImageNet and ImageNet-Subset.}
The various divisions reflect different properties. The 10/10 setting represents a long learning process while 50/5 indicates that the base of the learned knowledge is large, and the incremental learning process lasts for a long period. Previous studies only evaluated equally divided settings or the setting of learning half of the categories in the initial phase. In this study, a comprehensive evaluation of the proposed model is conducted.
The experimental results on CIFAR100 are shown in Table \ref{tab_avgacc_public_cifar100} and Table \ref{tab_reviewer2_Q5}, where the Ours-CNN(ont) uses ontology hierarchy and the Ours-CNN(vis) uses visual hierarchical clustering, with $k$ setting to 20 for CIFAR100 and 10 for miniImageNet.

\begin{table*}[htbp]
\centering
\caption{The last phase accuracy and average accuracy on CIFAR100.}
\label{tab_avgacc_public_cifar100}
\resizebox{0.7\textwidth}{!} {
\begin{tabular}{c|cccccccc}
\hline
\multirow{2}{*}{Methods} & \multicolumn{2}{c}{CIFAR100-10/10} & \multicolumn{2}{c}{CIFAR100-20/20} & \multicolumn{2}{c}{CIFAR100-50/5} & \multicolumn{2}{c}{CIFAR100-50/10} \\
 & last & Avg acc & last & Avg acc & last & Avg acc & last & Avg acc \\ \hline
iCaRL & 45.72 & 59.96 & 50.33 & 63.23 & 43.97 & 52.93 & 46.71 & 57.06 \\
LwF & 28.18 & 49.33 & 39.49 & 57.49 & 33.75 & 46.30 & 39.28 & 52.19 \\
LUCIR & 41.86 & 56.57 & 50.15 & 62.64 & 50.12 & 60.37 & 50.55 & 61.95 \\
Mnemonics & 42.42 & 58.53 & 48.95 & 63.18 & 50.79 & 60.43 & 53.58 & 63.12 \\
BiC & 47.30 & 57.56 & 40.90 & 48.33 & 41.63 & 54.11 & 50.88 & 58.35 \\
Ours-CNN(ont) & 45.81 & 61.02 & 56.03 & 66.45 & 52.59 & 61.50 & 58.20 & 65.45 \\
Ours-CNN(vis) & 47.72 & $\bm{62.38}$ & 57.00 & $\bm{68.27}$ & 52.69 & $\bm{61.51}$ & 58.63 & $\bm{65.69}$ \\ \hline
PODNet & 39.50 & 54.14 & 50.10 & 62.94 & 53.70 & 62.35 & 55.30 & 63.75 \\
Ours-PODNet & 40.70 & $\bm{54.61}$ & 50.70 & $\bm{63.44}$ & 53.90 & $\bm{62.85}$ & 55.10 & $\bm{64.32}$ \\ \hline
AANets-iCaRL & 45.32 & 61.40 & 51.75 & 65.40 & 47.91 & 59.79 & 50.24 & 61.71 \\
Ours-AANets & 45.63 & $\bm{63.02}$ & 49.42 & $\bm{65.72}$ & 47.33 & $\bm{59.93}$ & 50.26 & $\bm{62.19}$ \\ \hline
DER & 57.25 & 65.31 & 62.66 & 70.51 & 66.63 & 72.81 & 66.62 & 73.07 \\
Ours-DER & 58.03 & $\bm{65.42}$ & 62.94 & $\bm{70.74}$ & 66.61 & $\bm{73.06}$ & 66.82 & $\bm{73.27}$ \\ \hline
\end{tabular}
}
\end{table*}

\begin{table*}[htbp]
\centering
\caption{The last phase accuracy and average accuracy on miniImageNet.}
\label{tab_avgacc_public_miniImageNet}
\resizebox{0.8\textwidth}{!} {
\begin{tabular}{c|cccccccc}
\hline
\multirow{2}{*}{Methods} & \multicolumn{2}{c}{miniImageNet-10/10} & \multicolumn{2}{c}{miniImageNet-20/20} & \multicolumn{2}{c}{miniImageNet-50/5} & \multicolumn{2}{c}{miniImageNet-50/10} \\
 & last & Avg acc & last & Avg acc & last & Avg acc & last & Avg acc \\ \hline
iCaRL & 45.11 & 57.21 & 44.05 & 57.32 & - & - & 41.59 & 52.60 \\
LwF & 23.78 & 46.54 & 36.05 & 54.47 & 23.78 & 39.72 & 30.43 & 46.82 \\
LUCIR & 43.99 & 55.15 & 50.49 & 61.85 & 46.55 & 59.08 & 52.51 & 62.51 \\
BiC & 38.54 & 52.01 & 43.48 & 55.34 & 40.77 & 51.56 & 43.51 & 54.92 \\
GeoDL-CNN & 39.51 & 54.49 & 47.45 & 60.02 & 39.47 & 50.91 & - & - \\
Ours-CNN(ont) & 45.26 & $\bm{58.65}$ & 52.71 & 64.11 & 44.38 & 55.57 & 53.47 & 62.62 \\
Ours-CNN(vis) & 43.57 & 58.38 & 52.51 & $\bm{64.41}$ & 47.08 & $\bm{59.30}$ & 53.88 & $\bm{63.13}$ \\ \hline
PODNet & 38.50 & 53.69 & 47.50 & 60.96 & 43.30 & 54.45 & 47.10 & 59.37 \\
Ours-PODNet & 38.60 & $\bm{53.90}$ & 47.50 & $\bm{61.08}$ & 43.50 & $\bm{54.78}$ & 47.50 & $\bm{59.62}$ \\ \hline
AANets-iCaRL & 32.51 & 49.97 & 44.44 & 59.39 & 32.23 & 48.63 & 38.84 & 55.68 \\
Ours-AANets & 32.38 & $\bm{50.23}$ & 43.56 & $\bm{60.31}$ & 32.70 & $\bm{49.45}$ & 41.77 & $\bm{56.55}$ \\ \hline
DER & 56.99 & 65.53 & 61.10 & 69.02 & 59.04 & 65.45 & 59.96 & 66.49 \\
Ours-DER & 57.36 & $\bm{65.58}$ & 60.98 & $\bm{69.39}$ & 58.79 & $\bm{65.62}$ & 60.67 & $\bm{66.69}$ \\ \hline
\end{tabular}
}
\end{table*}

\begin{table*}[htbp]
    \centering
    \caption{Ablation experiment results under CIFAR100-10/10. CE is short for cross-entropy loss, CB is short for class-balanced classification loss, and KD is short for knowledge distillation. MG loss represents the multi-granularity loss. Decoupling represents retraining the classifier or not.}
    \label{tab_a1}
    \resizebox{1.0\textwidth}{!}{
        \begin{tabular}{ccccccccccccccccc}
        \hline
        Variations & CE & CB & KD & MG & Decoupling & init & 20 & 30 & 40 & 50 & 60 & 70 & 80 & 90 & 100 & Avg acc\\
        \hline
        baseline &\checkmark	&	&\checkmark	&	& 	& 90.40 & 65.40 & 52.40 & 42.55 & 38.74 & 34.78 & 33.03 & 29.03 & 28.18 & 26.80 & 44.13\\
        Ours-RW  &	&\checkmark	&\checkmark	&	&  & 89.20 & 68.15 & 52.50 & 42.63 & 38.90 & 34.30 & 33.50 & 28.79 & 28.90 & 27.19 & 44.41\\
        Ours-MGRW &	&\checkmark	&\checkmark	&\checkmark	&  & 91.70 & 68.45 & 53.93 & 45.38 & 39.74 & 36.35 & 35.23 & 31.60 & 30.64 & 29.29 & 46.23\\
        Ours-RS &\checkmark	&	&\checkmark	&	&\checkmark & 90.20 & 73.70 & 67.37 & 57.75 & 55.02 & 52.68 & 51.26 & 47.23 & 44.54 & 43.03 & 58.28\\
        Ours-MGRS &\checkmark	&	&\checkmark	&\checkmark &\checkmark	& 92.50 & 75.05 & 69.20 & 59.35 & 56.08 & 55.27 & 53.56 & 50.24 & 47.11 & 45.10 & 60.35\\
        Ours-RB &	&\checkmark	&\checkmark	&	&\checkmark	& 90.70 & 75.10 & 67.90 & 61.05 & 57.92 & 54.08 & 52.47 & 50.09 & 46.99 & 44.94 & 60.12\\
        Ours (MGRB) &	&\checkmark 	&\checkmark	&\checkmark &\checkmark	& 91.20 & $\bm{76.60}$ & $\bm{69.93}$ & $\bm{61.80}$ & $\bm{58.02}$ & $\bm{55.35}$ & $\bm{54.06}$ & $\bm{50.08}$ & $\bm{47.38}$ & $\bm{45.81}$ & $\bm{61.02}$\\
        \hline
        \end{tabular}
    }
\end{table*}

\begin{table}[h]
\centering
\renewcommand{\tablename}{Table}
\caption{Results of baseline methods against our method on CIFAR100-50/1.}
\label{tab_reviewer2_Q5}
\resizebox{0.25\textwidth}{!} {
\begin{tabular}{c|cc}
\hline
CIFAR100-50/1 & last & Avg acc \\ \hline
AANet-iCaRL & 43.78 & 55.01 \\
Ours-AANets & 44.09 & $\bm{55.24}$ \\ \hline
PODNet & 49.00 & 57.18 \\
Ours-PODNet & 49.80 & $\bm{57.42}$ \\ \hline
DER & 63.24 & 71.08 \\
Ours-DER & 63.52 & $\bm{71.37}$ \\ \hline
\end{tabular}}
\end{table}

\begin{table}[htbp]
\centering
\caption{The last phase accuracy and average accuracy on ImageNet-subset.}
\label{tab_avgacc_public_ImageNetSubset}
\resizebox{0.4\textwidth}{!}{
\begin{tabular}{c|cccc}
\hline
\multirow{2}{*}{Methods} & \multicolumn{2}{c}{ImageNet100-50/5} & \multicolumn{2}{c}{ImageNet100-50/10} \\
 & last & Avg acc & last & Avg acc \\
\hline
AANets-iCaRL & 39.89 & 55.25 & 54.65 & 65.58 \\
GeoDL-CNN & 45.02 & 57.80 & 50.72 & 62.07 \\
Ours-CNN(ont) & 49.36 & $\bm{62.68}$ & 59.62 & $\bm{69.75}$ \\ \hline
PODNet & 64.70 & 72.59 & 65.50 & 73.88 \\
Ours-PODNet & 64.90 & $\bm{72.77}$ & 65.90 & $\bm{74.13}$ \\ \hline
DER & 68.42 & 74.50 & 69.84 & 75.34 \\
Ours-DER & 68.58 & $\bm{74.62}$ & 70.24 & $\bm{75.61}$ \\ \hline
\end{tabular}
}
\end{table}

The proposed method outperforms PODNet by 8.24$\%$ and 5.33$\%$ under CIFAR100-10/10 and CIFAR100-20/20, respectively. It is worth noting that our method based on traditional CNN architectures can outperform many current CIL methods. We can also observe that our method can consistently improve baselines, including PODNet, AANets, and DER, without increasing the complexity of these models.

The experimental results of the miniImageNet are also shown in Table \ref{tab_avgacc_public_miniImageNet}. The value of GeoDL in miniImageNet-50/10 is vacant since the code reports an error. The observations on miniImageNet are consistent with the CIFAR datasets. Additionally, the average incremental accuracy of our method improves from 54.49$\%$ to 58.65$\%$ (+4.16$\%$) when compared with GeoDL under miniImageNet-10/10. It is evident that the proposed method can consistently improve the state-of-the-art methods without increasing the complexity of the models.

The experimental results of ImageNet-Subset in 50/5 and 50/10 settings are shown in Table \ref{tab_avgacc_public_ImageNetSubset}. 
The proposed method outperforms AANets-iCaRL by 4.17$\%$ and 7.43$\%$ under ImageNet-Subset-50/10 and ImageNet-Subset-50/5, respectively.

Under multiple partition rules and different datasets, the forgetting speed of our method is slower than that of the other methods, and we can further improve model performance on the state-of-the-art methods. This proves the effectiveness of our method.

\subsubsection{Ablation Studies on Public Datasets}
\label{sec:Experiments_ablation}

\textbf{The Experimental Setting.} \label{sec:Experiments_effect_component}
Our approach mainly comprises two components: re-balancing modeling and multi-granularity regularization. We first performed a few ablation experiments to analyze the contribution of each loss function. Then we compared the re-balancing modeling and multi-granularity regularization components based on the above experimental results. Note that we used the ontology for building the hierarchy. Experimental results are shown in Table \ref{tab_a1}.

To verify the contribution of each loss function,  we conducted the following experiments.
\begin{itemize}

    \item \noindent\textbf{Baseline} method has been introduced in Section \ref{sec:Method_Baseline}. It uses the cross-entropy loss function on all classes and knowledge distillation loss on old classes.

    \item \noindent\textbf{Ours-RW} applies the re-weighting method to deal with data imbalance.

    \item \noindent\textbf{Ours-MGRW} applies the multi-granularity regularized re-weighting modeling and does not apply the balanced set to the retraining stage.

    \item \noindent\textbf{Ours-RS} applies the re-sampling method to alleviate data imbalance.

    \item \noindent\textbf{Ours-MGRS} trains the model using multi-granularity regularized re-sampling modeling, including the cross-entropy classification loss, knowledge distillation loss as well as multi-granularity loss function.
    The difference between Ours-MGRS and Ours-MGRB is that Ours-MGRB has an additional re-weighting component to further reduce the imbalance.

    \item \noindent\textbf{Ours-RB} uses re-balancing modeling, including re-weighting and re-sampling. The loss function is composed of the class-balanced classification loss function and knowledge distillation loss.

    \item \noindent\textbf{Ours (MGRB)} employs the class-balanced classification loss function, knowledge distillation loss, and multi-granularity loss. Compared to the Ours-RB, a multi-granularity regularization term is added.
\end{itemize}

The results of the ablation experiments are listed in Table \ref{tab_a1}. The ablation analysis of re-balancing modeling and multi-granularity regularization is illustrated below.

\begin{figure}[htb]
    \centering
    \includegraphics[width=0.48\textwidth]{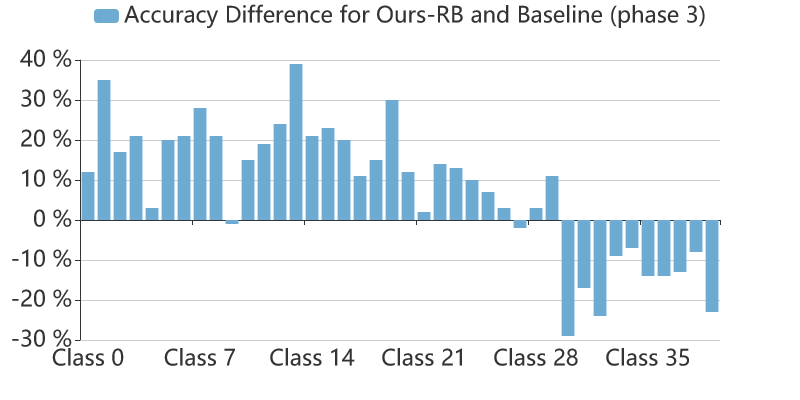}
    \caption{The accuracy difference of CIFAR100-10/10 in phase 3 between Ours-RB method and Baseline method. Class 0-29 are old classes learned before, and the last 10 classes are newly coming classes.}
    \label{fig_acc_differenece_phase4}
\end{figure}

\begin{figure}[htb]
    \centering
    \includegraphics[width=0.48\textwidth]{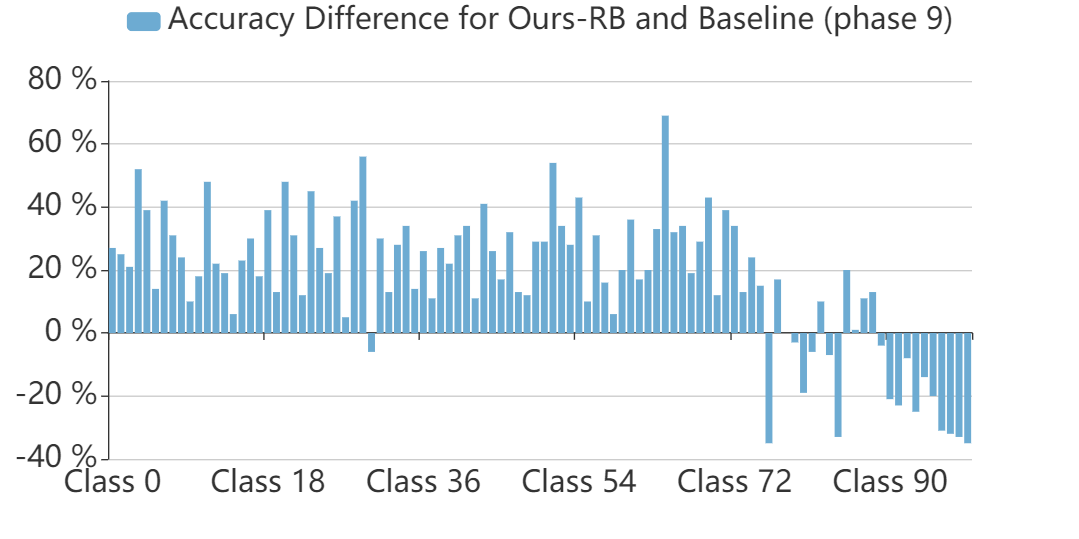}
    \caption{The accuracy difference of CIFAR100-10/10 in phase 9 between Ours-RB method and Baseline method. Class 0-89 are old classes learned before, and Class 90-99 are newly coming classes.}
    \label{fig_acc_differenece_phase9}
\end{figure}

\begin{figure}[htb]
    \centering
    \includegraphics[width=0.48\textwidth]{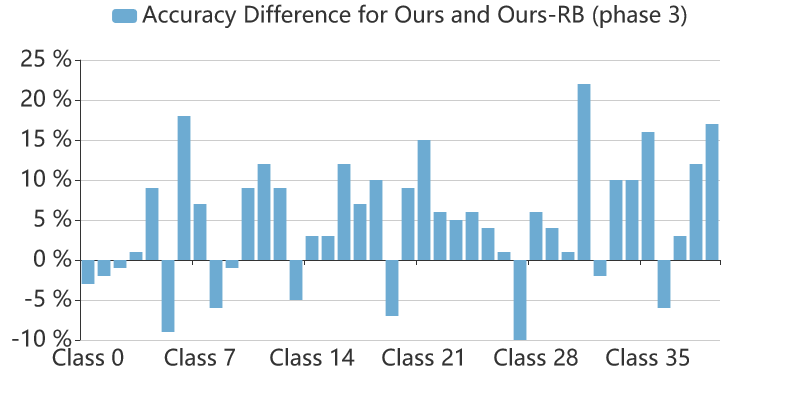}
    \caption{The accuracy difference of CIFAR100-10/10 in phase 3 between Ours and Ours-RB method. Class 0-29 are old classes learned before, and the last 10 classes are newly coming classes.}
    \label{fig_acc_differenece_phase3_cmp}
\end{figure}

\textbf{The Effect of Re-balancing Modeling.} \label{sec:Experiments_effect_decoupling}
In this section, we analyze the effects of re-balancing modeling. 
\textbf{Ours-RS} adds the retraining stage, and it shows a significant increase in accuracy. When adding re-weighting and re-sampling components separately, we observed a 0.28$\%$ and 14.15$\%$ increase in accuracy, indicating significant biases in the classifier, and the re-sampling strategy plays a vital role in re-balancing modeling.

Moreover, when the class-balanced classification loss (re-weighting method) is used (compare \textbf{Ours-MGRS} and \textbf{Ours (MGRB)}), the accuracy is 0.67$\%$ higher than that of learning with the cross-entropy loss function. Because the samples are extremely imbalanced, old classes are weakly supervised during the learning process. The class-balanced classification loss re-balances the penalties of new and old classes to ensure superior results. 

Compared \textbf{Ours-RB} with \textbf{Baseline}, the re-balancing method increases the average accuracy by 15.99$\%$. We further output the accuracy of each class, and subtract the accuracy of each class in \textbf{Baseline} with \textbf{Ours-RB} in phase 3 and phase 9, as shown in Fig. \ref{fig_acc_differenece_phase4} and Fig. \ref{fig_acc_differenece_phase9}. It can be observed that the accuracy of the newly learned classes decreased significantly. This may be owing to the under-fitting of the new classes caused by the weakening of their weights.

Ablation experiments with other settings on the CIFAR100 dataset can be found in Table \ref{tab_abl_c100other}.
\begin{table}[htbp]
\centering
\renewcommand{\tablename}{Table}
\caption{Ablation experiment results under CIFAR100-20/20, CIFAR100-50/5 and CIFAR100-50/10.}
\label{tab_abl_c100other}
\resizebox{0.5\textwidth}{!} {
\begin{tabular}{c|cccccc}
\hline
\multirow{2}{*}{Methods} & \multicolumn{2}{c}{CIFAR100-20/20} & \multicolumn{2}{c}{CIFAR100-50/5} & \multicolumn{2}{c}{CIFAR100-50/10} \\
 & Last & Avg acc & Last & Avg acc & Last & Avg acc \\ \hline
baseline & 34.33 & 52.79 & 32.88 & 42.64 & 34.87 & 47.85 \\
Ours-RW & 35.35 & 53.76 & 36.30 & 46.78 & 35.95 & 49.99 \\
Ours-MGRW & 38.02 & 55.15 & 37.06 & 47.48 & 35.05 & 50.35 \\
Ours-RS & 53.35 & 64.14 & 45.53 & 56.68 & 55.67 & 63.84 \\
Ours-MGRS & 55.80 & 66.33 & 51.37 & 60.30 & 57.42 & 65.12 \\
Ours-RB & 54.32 & 65.69 & 51.32 & 60.48 & 57.27 & 64.62 \\
Ours(MGRB) & 56.03 & $\bm{66.45}$ & 52.59 & $\bm{61.50}$ & 58.20 & $\bm{65.45}$ \\ \hline
\end{tabular}}
\end{table}

\begin{table}[htbp]
\centering
\renewcommand{\tablename}{Table}
\caption{The impact without $L_{CE}$ and $L_{CB}$ on CIFAR100-50/10.}
\label{tab_reviewer2_Q4}
\resizebox{0.48\textwidth}{!} {
\begin{tabular}{cccccccc}
\hline
CIFAR100-50/10 & 50 & 60 & 70 & 80 & 90 & 100 & Avg acc \\ \hline
w/o L\_CE/L\_CB & 77.50 & 70.18 & 64.83 & 61.95 & 58.01 & 57.18 & 64.94\\
Ours & 77.38 & 70.18 & 65.60 & 62.43 & 58.91 & 58.20 & $\bm{65.45}$ \\
\hline
\end{tabular}}
\end{table}
\begin{table}[htbp]
\centering
\renewcommand{\tablename}{Table}
\caption{The impact without $L_{CE}$ and $L_{CB}$ on miniImageNet-20/20.}
\label{tab_reviewer2_Q4_2}
\resizebox{0.48\textwidth}{!} {
\begin{tabular}{ccccccc}
\hline
MiniImageNet-20/20 & 20 & 40 & 60 & 80 & 100 & Avg acc \\
\hline
w/o L\_CE/L\_CB & 78.20 & 64.65 & 58.47 & 53.66 & 49.65 & 60.93 \\
Ours & 79.25 & 68.80 & 62.05& 57.73 & 52.71& $\bm{64.11}$ \\\hline
\end{tabular}}
\end{table}
\begin{table}[htbp]
\centering
\renewcommand{\tablename}{Table}
\caption{The impact of $L_H$ on joint and class incremental learning setting.}
\label{tab_reviewer2_Q2}
\resizebox{0.4\textwidth}{!} {
\begin{tabular}{ll|cc}
\hline
\multicolumn{2}{l|}{Setting} & w/o L\_H & w/ L\_H \\ \hline
\multicolumn{1}{l|}{\multirow{2}{*}{CIFAR100-50/10}} & Joint & 72.03 & 72.66 \\
\multicolumn{1}{l|}{} & CIL & 64.62 & 65.46 \\ \hline
\multicolumn{1}{l|}{\multirow{2}{*}{miniImageNet-20/20}} & Joint & 64.59 & 65.26 \\
\multicolumn{1}{l|}{} & CIL & 63.03 & 64.11 \\ \hline
\end{tabular} }
\end{table}

\begin{table}[htbp]
\centering
\caption{The influence of hierarchical structure on the accuracy of CIFAR100-20/20. The clustering parameter $k$ is set to 20.}
\label{tab_a2}
\begin{tabular}{ccccccc}
\hline
Method	&init	&40	&60 &80	&100 & Avg acc\\
\hline
NMG	    &82.90	&69.93	&62.88	&58.43	&54.32  &65.69\\
MG(ont)	&82.60  &70.35  &63.92  &59.59  &57.06  &66.70\\
MG(sem)	&82.65 	&71.05 	&64.25 	&60.31 	&57.08 	&67.07\\
MG(vis)	&84.55  &72.20  &66.75 	&60.84 	&57.00 	&$\bm{68.27}$
\\
\hline
\end{tabular}
\end{table}

\begin{figure}[ht]
    \centering
    \includegraphics[width=0.48\textwidth]{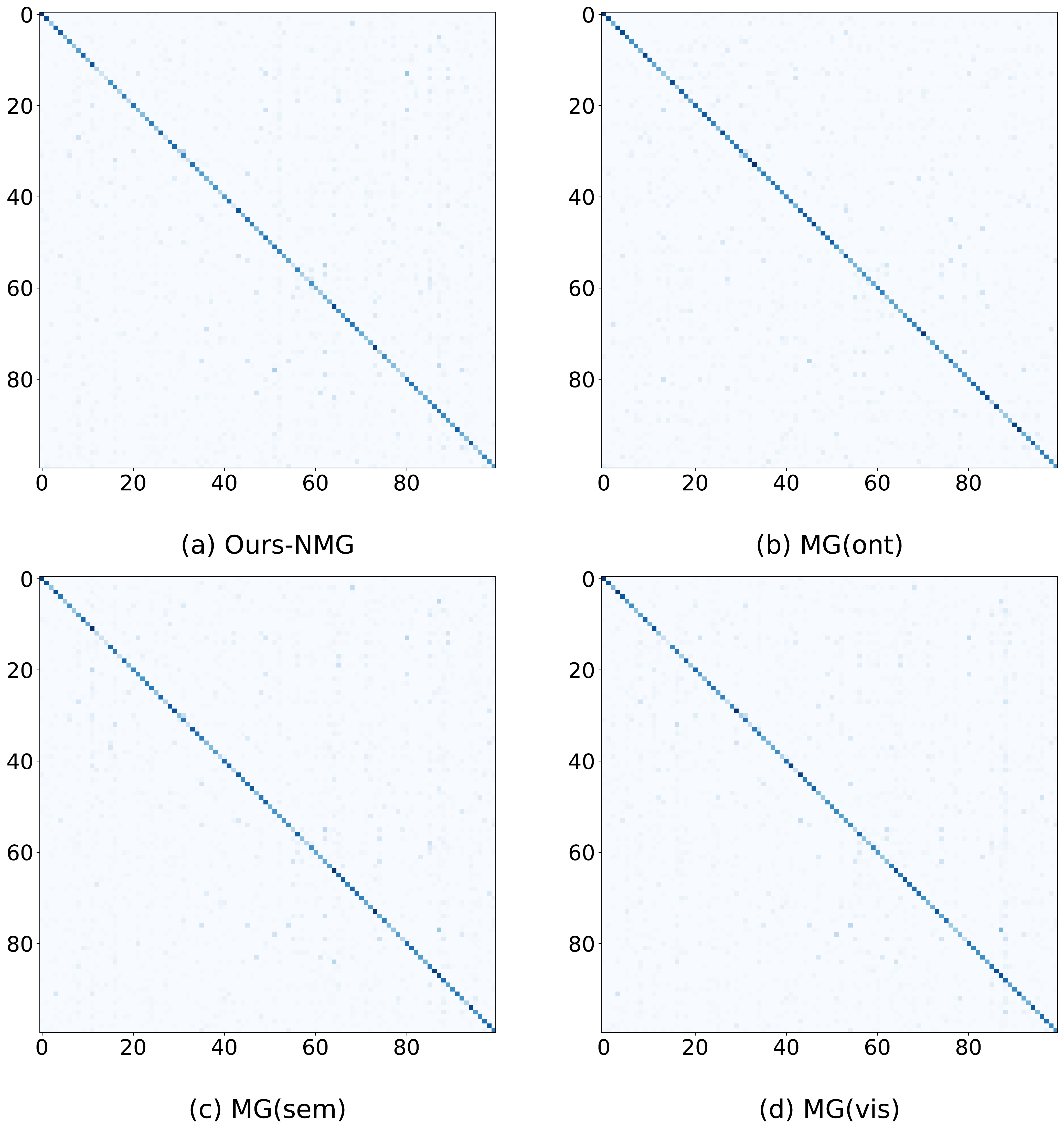}
    \caption{Confusion matrices of four variations on CIFAR100-20/20 from 80 classes to 100 classes: (a)NMG: do not use multi-granularity regularization; (b) MG(ont): use multi-granularity regularization and build hierarchy by using the ontology of CIFAR100; (c) MG(sem): use multi-granularity regularization and build hierarchy by semantic clustering and (d) MG(vis): use multi-granularity regularization and build hierarchy by visual hierarchical clustering.}
    \label{CIFAR100-20/20_CM}
\end{figure}

\begin{figure}[htp]
    \centering
    \includegraphics[width=0.48\textwidth]{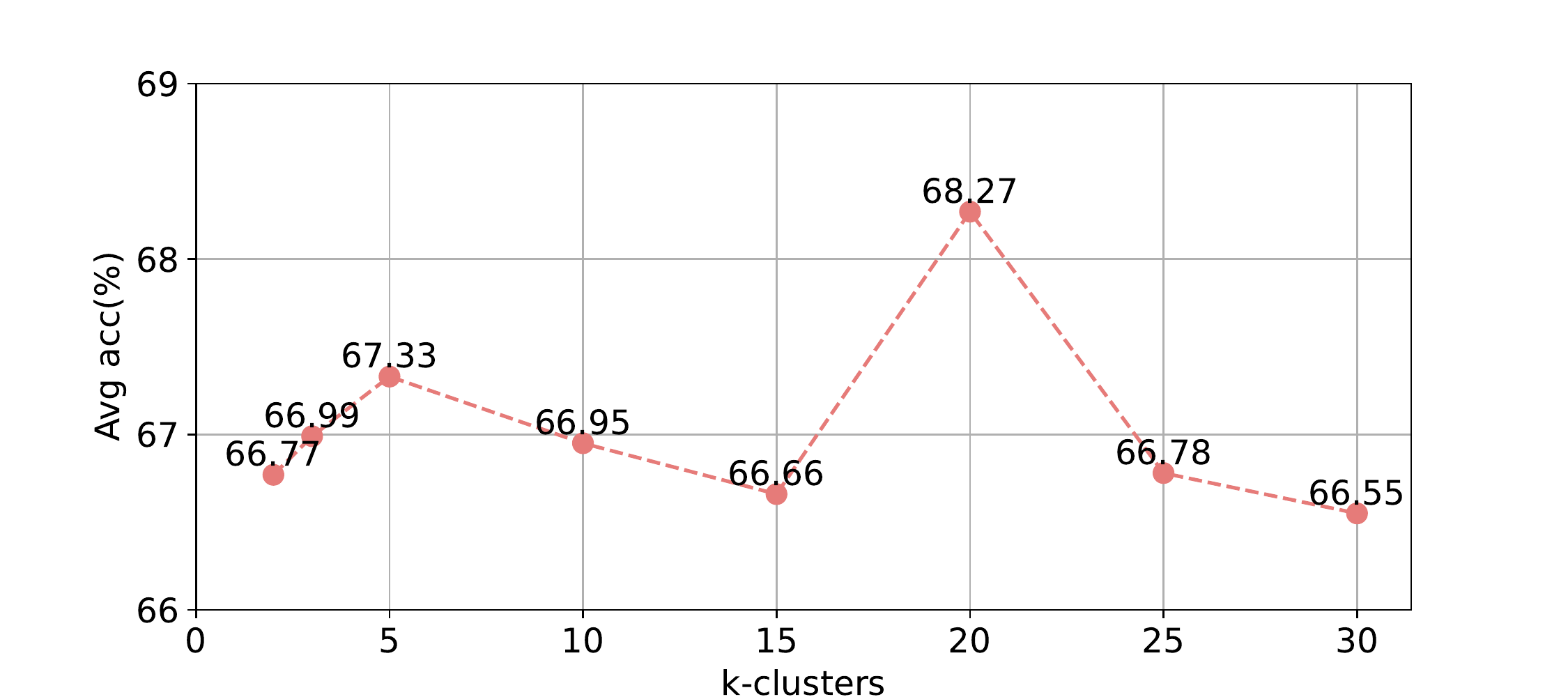}
    \caption{Accuracy on CIFAR100-20/20 using $k$ visual hierarchy clustering. When $k=20$, incremental learning achieves the highest average accuracy.}
    \label{fig_kcluster}
\end{figure}

\begin{figure*}[htb]
    \centering
    \includegraphics[width=\textwidth]{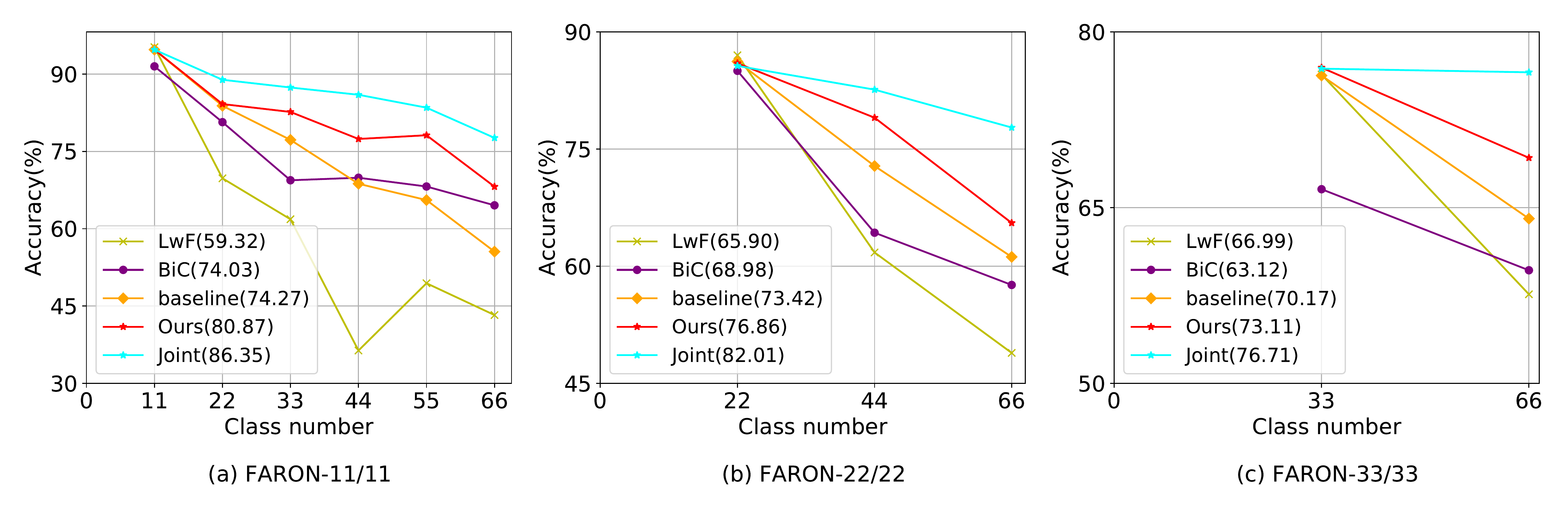}
    \caption{Incremental learning results (accuracy) on FARON with settings of (a)11/11: 11 classes of samples are presented initially and are given in each incremental phase, (b) 22/22, (c) 33/33. The average accuracy is noted in the legend.}
    \label{fig_FARON}
\end{figure*}

We also conducted experiments without $L_{CE}$ and $L_{CB}$ on CIFAR100-50/10 and miniImageNet-20/20 to explore the impact of $L_{CE}$/$L_{CB}$. The results are shown in Table \ref{tab_reviewer2_Q4} and Table \ref{tab_reviewer2_Q4_2}. We can observe that``Ours'' are better than ``without $L_{CE}$/$L_{CB}$'', which indicates a significant classification loss. The reason may be that the multi-granularity regularization provides supplemental information for classification objectives but cannot replace this objective. Moreover, the proposed method is fundamentally different from label smoothing despite somewhat similar. Label smoothing treats all classes as the same, whereas the essence of multi-granularity regularization is to introduce correlations between classes to improve performance.

\textbf{The Effect of Multi-granularity Regularization.} \label{sec:Experiments_effect_hierarchy}
When comparing \textbf{Ours-RB} and \textbf{Ours (MGRB)}, it is observed that when the multi-granularity loss is used, the accuracy of all phases improves. Specifically, the final accuracy rate increased from 44.94$\%$ to 45.81$\%$, and the average accuracy rate increased by approximately 0.9$\%$. We further output the accuracy difference of each class between the \textbf{Ours (MGRB)} and \textbf{Ours-RB}, as shown in Fig. \ref{fig_acc_differenece_phase3_cmp}. It can be observed that multi-granularity regularization is no longer susceptible to the under-fitting of new classes. The relationship contained in the label is consequently helpful in strengthening the learning of both old and new classes.

To discuss the impact of $L_H$ on the joint model and an incremental model, we conducted experiments in two settings: (1) all classes are trained at once, and an investigation into how much influence $L_H$ on the ``Joint" baseline is done by removing or adding the multi-granularity regularization term; (2) We followed the class incremental learning settings and tested the impact of $L_H$ by removing or adding the multi-granularity regularization term. 
Experimental results are shown in Table \ref{tab_reviewer2_Q2}. It can be seen that the multi-granularity regularization assists more in the class incremental scenario than in the raw joint model, demonstrating that it can facilitate the learning of incremental learning models.

In addition to exploring the effectiveness of multi-granularity regularization, the impact of modeling methods is verified. We conducted the following experiments to compare three modeling methods.
\begin{itemize}
    \item \noindent\textbf{NMG} does not use multi-granularity regularization.

    \item \noindent\textbf{MG(ont)} uses multi-granularity regularization based on ontology.
    
    \item \noindent\textbf{MG(sem)} employs multi-granularity regularization based on semantic hierarchical clustering.
    
    \item \noindent\textbf{MG(vis)} employs multi-granularity regularization based on visual hierarchical clustering.
    
\end{itemize}

The experimental results of the above experiments are shown in Table \ref{tab_a2}. We also drew the confusion matrices of these four settings in Fig. \ref{CIFAR100-20/20_CM}. It can be observed that the performance of old classes is consolidated and the performance of the new classes is enhanced.

As for the visual hierarchical clustering, we evaluated the average accuracy on CIFAR100-20/20 with multiple $k$. The results are presented in Fig. \ref{fig_kcluster}. As the number of clusters continues to increase, the average accuracy fluctuates. 
This shows that the choice of the number of visual clustering needs a balance. Various clustering numbers result in different relationships among classes, and the ability to prevent forgetting is different.


\subsection{Experiments on Fault Diagnosis}

\subsubsection{Experimental Setups}
\label{sec:FD_Experimental_setup}

\begin{figure}[ht]
    \centering
    \includegraphics[width=0.48\textwidth]{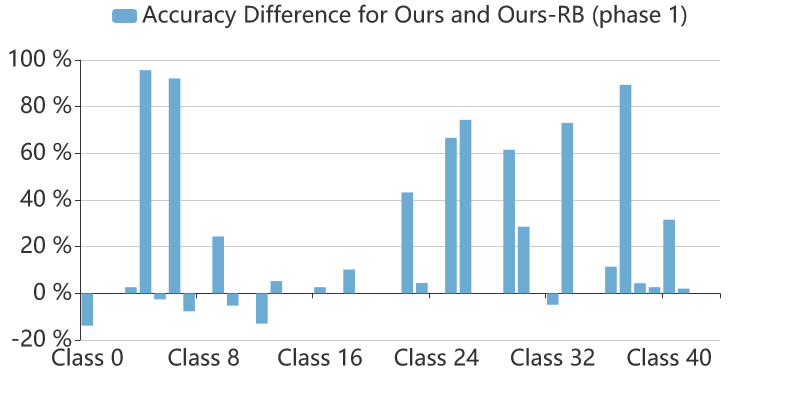}
    \caption{The accuracy difference of FASON-22/22 in phase 1 between Ours and Ours-RB method. Class 0-21 are old classes learned before, and the last 22 classes are newly coming classes.}
    \label{fig_acc_differenece_phase1_cmp_FASON}
\end{figure}

\begin{figure}[ht]
    \centering
    \includegraphics[width=0.48\textwidth]{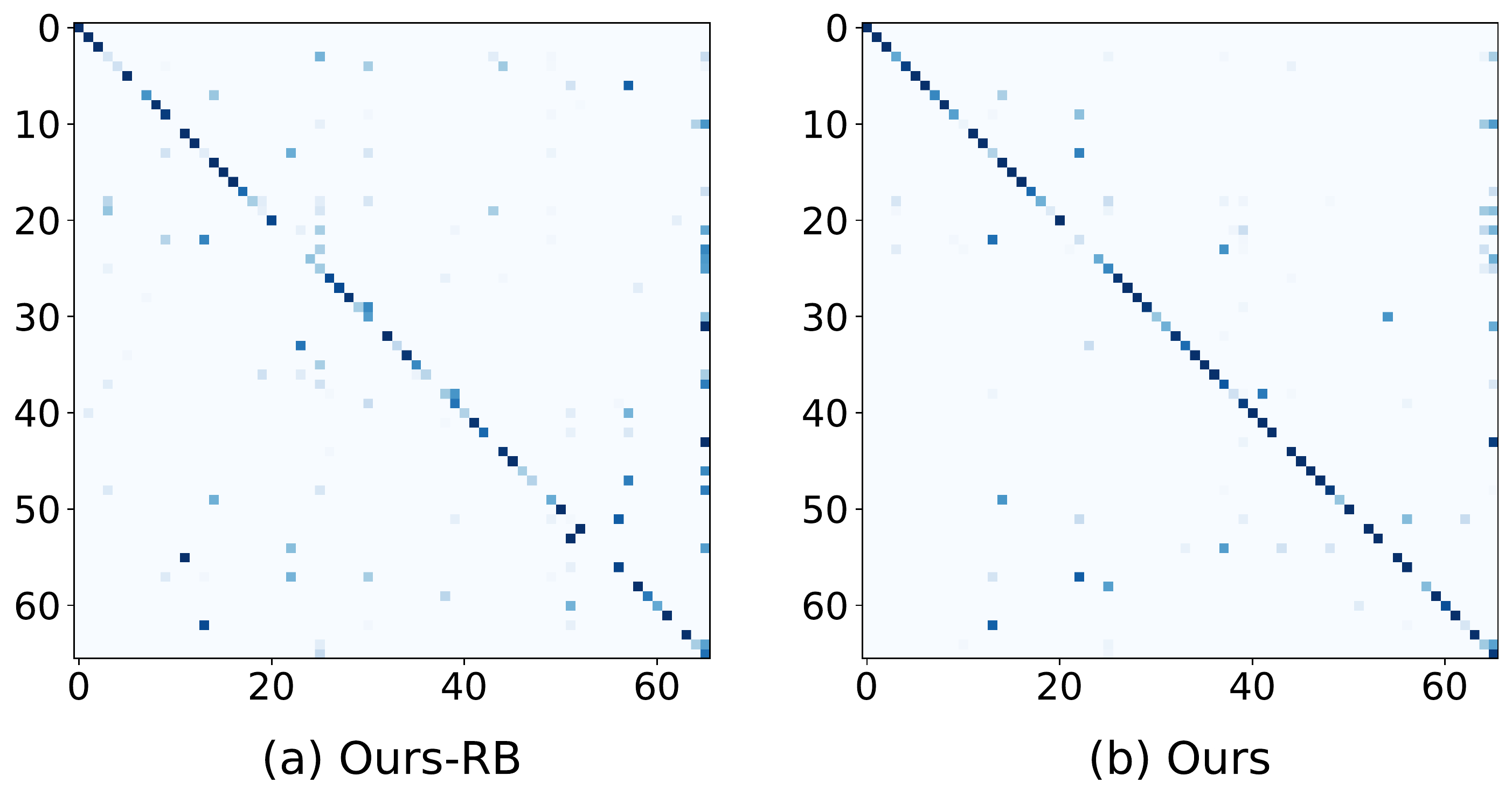}
    \caption{Confusion matrices of Ours-RB and Ours on FASON-22/22 from 44 classes to 66 classes. (a) Ours-RB only uses re-balancing modeling, and (b) Ours uses the multi-granularity regularized re-balancing method.}
    \label{fig_FASON_CM}
\end{figure}

\begin{figure}[ht]
    \centering
    \includegraphics[width=0.48\textwidth]{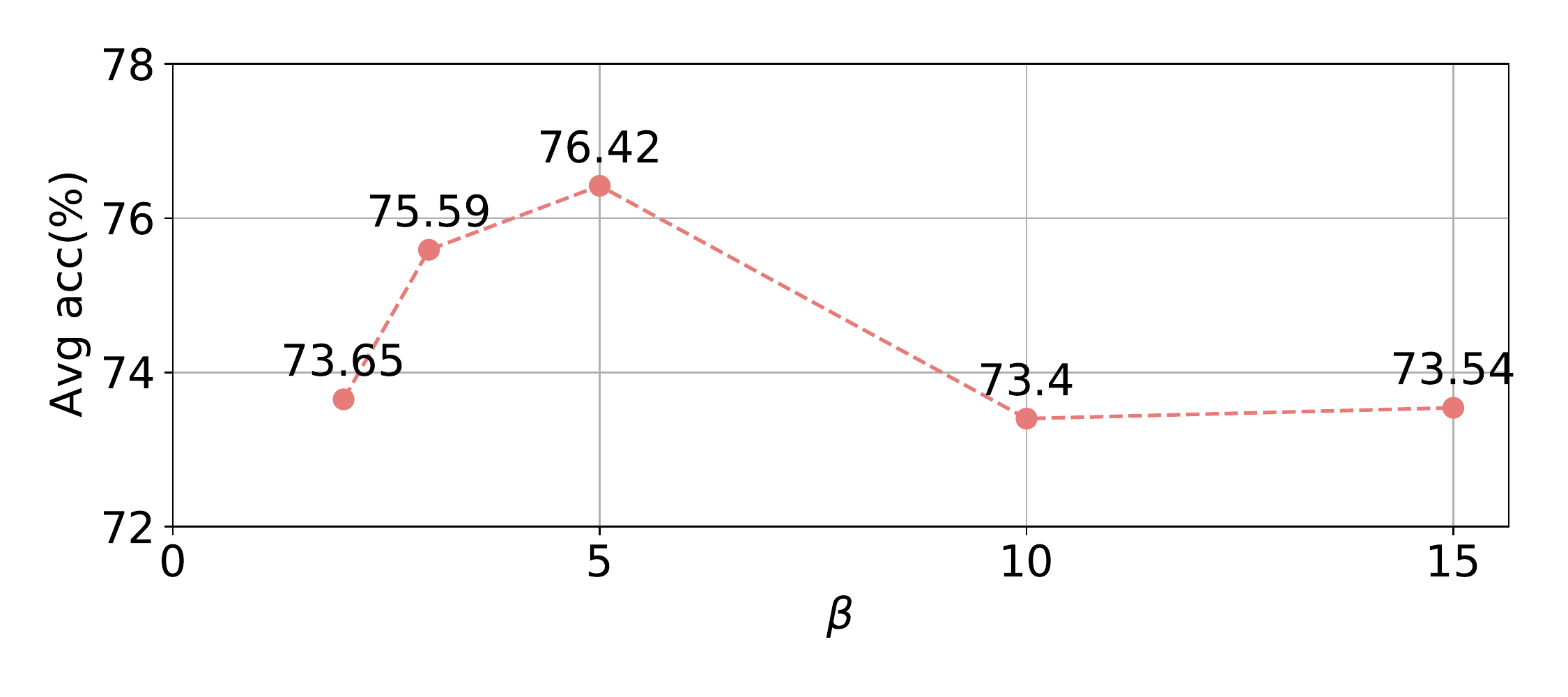}
    \caption{The accuracy of FARON-22/22 when using different $\beta$ to control the label distribution.}
    \label{fig_FASON_beta}
\end{figure}

\begin{figure}[ht]
    \centering
    \includegraphics[width=0.48\textwidth]{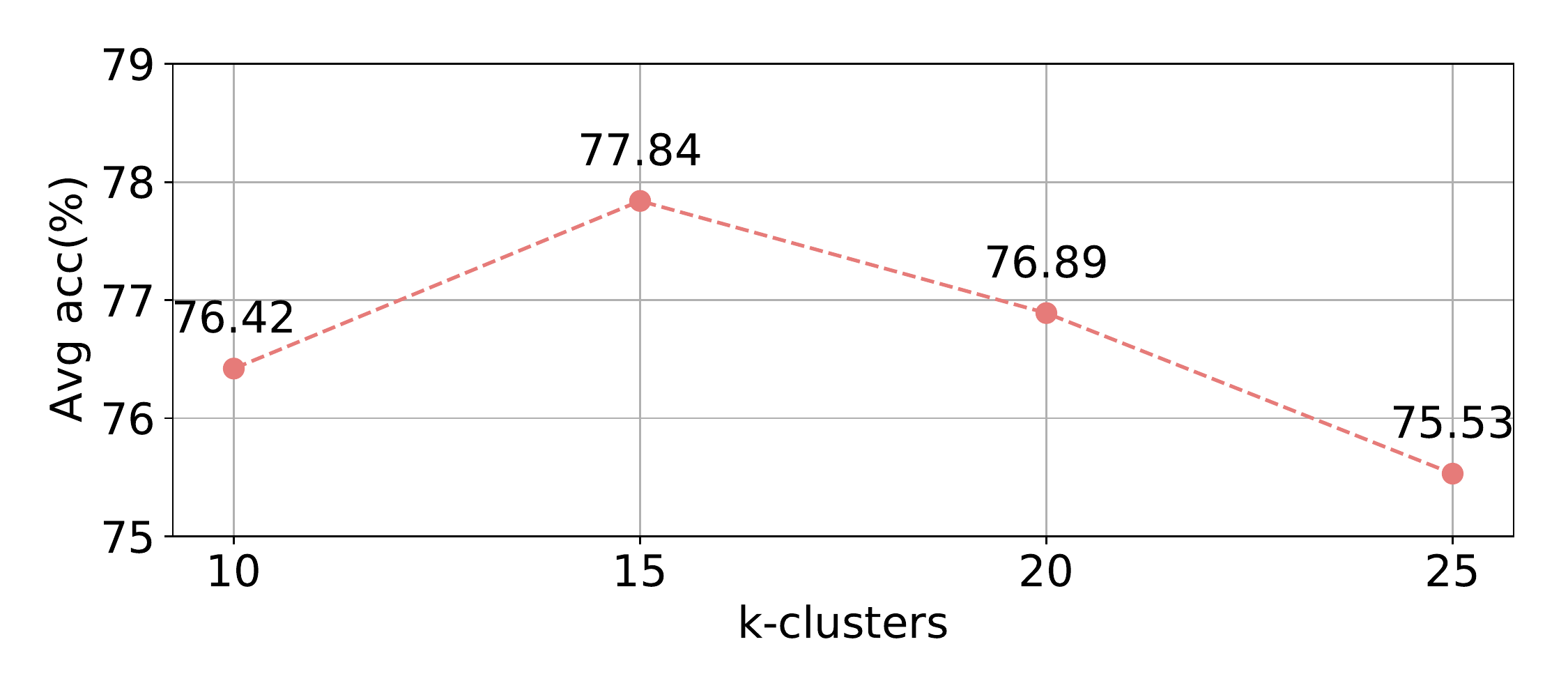}
    \caption{The Accuracy on FARON-22/22 grouping into various clusters, reflecting the change in accuracy caused by the change of the number of cluster.}
    \label{fig_FASON_k-cluster}
\end{figure}

Fault Recognition Of Nuclear power system (FARON), proposed by Wang et al.~\cite{wang2021coarse}, is a large-scale fault diagnosis dataset collected for a real-world scenario. We adopt this dataset to explore the effects of the proposed method on real-world data.
This dataset is generated from a complex nuclear power system. The outputs are the data of 121 sensors in the system under different conditions. It contains 65 kinds of fault types and one normal type. The length of sampling time is 20 minutes. The dimension of each sample is 20 $\times$ 121 after data preprocessing, and the entire dataset contains 12,643 samples for training and 3,562 samples for testing. Among the 66 classes, each fault type has a different number of samples, at least 83 samples and at most 2,681 samples. Because other methods do not provide codes that could be applied to the fault diagnosis dataset, we extend the baseline method in LwF\cite{li2018learning}, BiC\cite{wu2019large}, and employed them to compare with our method.

\begin{table}[!ht]
    \centering
    \caption{The ablation study on FARON. The initial 22 classes are trained first, and then 22 new classes are added in each phase.}
    \label{tab_FARON_cmp}
    \begin{tabular}{ccccc}
    \hline
    Method & init & 44 & 66 & Avg acc \\
    \hline
    baseline  & 86.22 &	72.84 &	61.20 &	73.42\\
    Ours-RW   & 85.83 &	73.48 &	61.79 &	73.70\\
    Ours-MGRW & 86.00 &	79.34 & 62.41 & 75.91\\
    Ours-RS   & 85.88 &	75.03 &	58.42 &	73.11\\
    Ours-MGRS & 84.64 & 78.96 & 65.67 & 76.42\\
    Ours-RB   & 86.51 & 74.35 &	60.92 &	73.93\\
    Ours (MGRB)& 86.00 & 79.03 & 65.55 & $\bm{76.86}$\\
    \hline
    \end{tabular}
\end{table}

\subsubsection{Implementation Details}
\label{sec:Implementation_detail_FARON}

For FARON, we adopted the backbone network proposed by Wang et al.~\cite{wang2021coarse}. We used an Adam optimizer and set the learning rate to 1e-5. The learning order was generated by setting the seed to 36, thus making the normal class appear in the initial phase. The number of epochs for the imbalance training and retraining stages were set to 50 and 10, respectively. The maximum memory was set to $K=8,000$.
The fault diagnosis dataset was split to assess 11/11, 22/22, and 33/33. $\beta$ was set to 5 in all cases. For the scale of the loss term, we set $\alpha$ to 1.0, 0.1, and 0.1, respectively. The hierarchy of FARON was built by clustering the mean of all samples in each class and grouping them into 15 coarse categories.

\subsubsection{Results on Fault Diagnosis}
\label{sec:Experiments_evaluation_FARON}

The experimental results on FARON are shown in Fig. \ref{fig_FARON}. In FARON-33/33, our method achieves a 9.60$\%$ improvement in the final phase and a 9.99$\%$ improvement on the average accuracy compared with BiC. This is because the dataset is derived from a complex nuclear power system, and the biases are more complicated. Experimental results verify that the proposed method can be applied to complex and real scenarios.

\subsubsection{Ablation Studies on Fault Diagnosis}
\label{sec:Experiments_ablation_FARON}

Our method achieves such a significant improvement owing to the maintenance of the category hierarchy. It can be observed from Table \ref{tab_FARON_cmp} that the multi-granularity loss function has an obvious effect on this dataset. As shown in Fig. \ref{fig_acc_differenece_phase1_cmp_FASON}, the difference in accuracy between \textbf{Ours (MGRB)} and \textbf{Ours-RB} further reflects the necessity of multi-granularity regularization. From Fig. \ref{fig_FASON_CM}, it can be observed more directly that multi-granularity can reduce the confusion of both old classes and new classes.

To analyze the impact of the multi-granularity regularization term on the accuracy of the overall model, we conduct experiments on FARON-22/22 with $k=10, 15, 20, 25$ and $\beta=2,3,5,10,15$. The experimental results are shown in Fig. \ref{fig_FASON_beta} and Fig. \ref{fig_FASON_k-cluster}. It is found that setting $\beta$ to 5 leads to better results. The experimental results indicate a trade-off between excessively uniform and sharp label distribution. 

\section{Conclusion}
\label{sec:Conclusion}
In this paper, we proposed a new multi-granularity regularized re-balancing method to address catastrophic forgetting. We first used re-balancing modeling to reduce the effect of data imbalance. We found that the re-balancing strategy can produce an assumption-agnostic approach to reduce the huge biases of the models. However, they would under-fit the new classes and lead to performance degradation. Therefore, we further designed a multi-granularity regularization term to embed the correlations of classes by optimizing a continuous label distribution target based on the knowledge hierarchy, which was built according to ontology or semantic hierarchical clustering. In this way, the under-fitting problem is eliminated, and the learning of both old and new classes can be boosted simultaneously.
We conducted extensive experiments on public datasets and a fault diagnosis dataset. Results show that the proposed method could achieve superior performance compared to classical and the latest methods. Typically, it can obtain up to 7.47$\%$ and 9.99$\%$ improvement on average accuracy on various CIL settings, respectively.



\ifCLASSOPTIONcaptionsoff
  \newpage
\fi



\bibliographystyle{IEEEtran}
\bibliography{bare_jrnl}

\vspace{-10mm}

\begin{IEEEbiography}[{\includegraphics[width=1in,height=1.25in,clip,keepaspectratio]{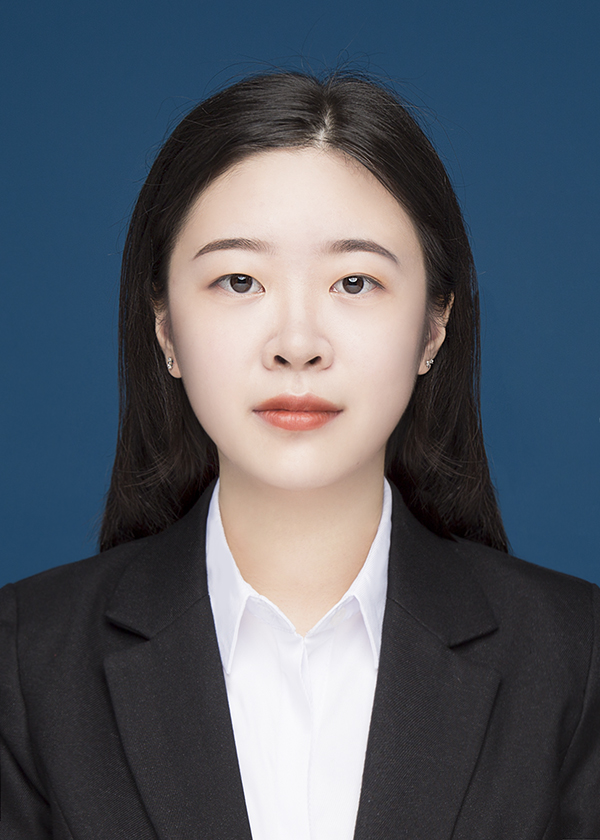}}]{Huitong Chen} received the B.S. degree in computer science and technology from Tianjin University in 2021. She is currently pursuing the graduation degree in computer science and technology with the College of Intelligence and Computing, Tianjin University, Tianjin, China. Her research is focused on incremental learning, data mining, and machine learning.

\end{IEEEbiography}

\vspace{-10mm}

\begin{IEEEbiography}[{\includegraphics[width=1in,height=1.25in,clip,keepaspectratio]{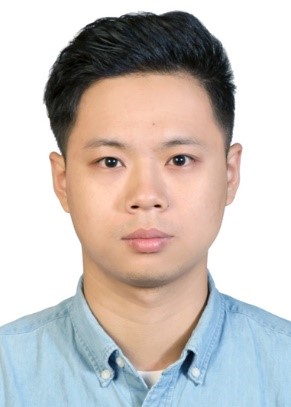}}]{Yu Wang} received the B.S. degree in communication engineering, the M.S. degree in software engineering, and Ph.D. degree in computer applications and techniques from Tianjin University in 2013 and 2016, and 2020, respectively. He is currently an assistant professor of Tianjin University. His research is focused on data mining and machine learning, especially multi-granularity learning in open and dynamic environment for computer vision and industrial applications.

\end{IEEEbiography}

\vspace{-10mm}

\begin{IEEEbiography}[{\includegraphics[width=1in,height=1.25in,clip,keepaspectratio]{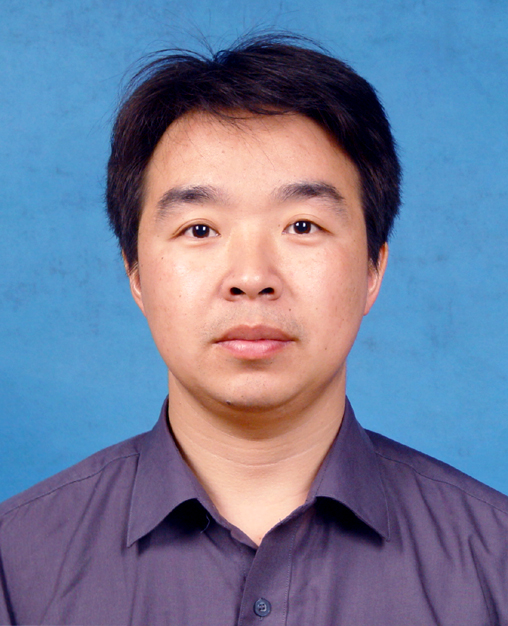}}]{Qinghua Hu} received the B.S., M.S., and Ph.D. degrees from the Harbin Institute of Technology, Harbin, China, in 1999, 2002, and 2008, respectively. He has published over 200 peer-reviewed papers. His current research is focused on uncertainty modeling in big data, machine learning with multi-modality data, intelligent unmanned systems. He is an Associate Editor of the IEEE TRANSACTIONS ON FUZZY SYSTEMS, Acta Automatica Sinica, and Energies.
\end{IEEEbiography}
%



%



\end{document}